\documentclass[pdflatex,sn-mathphys-num]{lucid}

\usepackage{graphicx}%
\usepackage{multirow}%
\usepackage{amsmath,amssymb,amsfonts}%
\usepackage{amsthm}%
\usepackage{mathrsfs}%
\usepackage[title]{appendix}%
\usepackage{xcolor}%
\usepackage{textcomp}%
\usepackage{manyfoot}%
\usepackage{booktabs}%
\usepackage{array}%
\usepackage{tabularx}%
\usepackage{algorithm}%
\usepackage{algorithmicx}%
\usepackage{algpseudocode}%
\usepackage{listings}%
\usepackage{graphicx}
\usepackage{newfloat}
\supplement{supplementary}

\DeclareFloatingEnvironment[
  name={Extended Data Fig.},
  autorefname={Extended Data Fig.},
  listname={List of Extended Data Figures},
  fileext=loedf,
  placement=htbp,
  within=none
]{extendedfigure}

\DeclareFloatingEnvironment[
  name={Extended Data Table},
  autorefname={Extended Data Table},
  listname={List of Extended Data Tables},
  fileext=loedt,
  placement=htbp,
  within=none
]{extendedtable}

\makeatletter
\def\EDFigName{extendedfigure}
\long\def\@makecaption#1#2{%
  \ifx\FigName\@captype\vskip\abovecaptionskip\@figurecaption{#1}{#2}%
  \else\ifx\EDFigName\@captype\vskip\abovecaptionskip\@figurecaption{#1}{#2}%
  \else\@tablecaption{#1}{#2}\vskip\belowcaptionskip\fi\fi}
\long\def\@tablecaption#1#2{%
  \setbox\tabcapbox\vbox{\tablecaptionfont{\bfseries #1}{\hskip2mm}#2\vphantom{y}\par}%
  \box\tabcapbox}
\makeatother

\theoremstyle{thmstyleone}%
\theoremstyle{thmstyletwo}%

\theoremstyle{thmstylethree}%

\begin{document}

\title[Article Title]{Naming places makes AI models fabricate disaster damage}


\author[1,2]{\fnm{Hongruixuan} \sur{Chen}}\email{qschrx@gmail.com}
\author[1,2]{\fnm{Pengyu} \sur{Dai}}
\author[3]{\fnm{Haolin} \sur{Wang}}
\author[4]{\fnm{Li} \sur{Mi}}
\author[1]{\fnm{Yafei} \sur{Ou}}
\author[4]{\fnm{Olivier} \sur{Dietrich}}
\author[1,2]{\fnm{Jian} \sur{Song}}
\author[1,2]{\fnm{Dong-Dong} \sur{Wu}}
\author[2]{\fnm{Yuzeng} \sur{Chen}}
\author[1]{\fnm{Xuanchao} \sur{Fu}}
\author[2]{\fnm{Junjue} \sur{Wang}}
\author[5]{\fnm{Hao} \sur{Li}}
\author[1]{\fnm{Junshi} \sur{Xia}}
\author[1,2]{\fnm{Takashi} \sur{Ishida}}
\author[6,7]{\fnm{Valentin} \sur{Hofmann}}
\author[8]{\fnm{Jocelyn} \sur{Chanussot}}
\author[9]{\fnm{Zhe} \sur{Zhu}}
\author[1,2]{\fnm{Masashi} \sur{Sugiyama}}
\author[4]{\fnm{Konrad} \sur{Schindler}}
\author[1,2]{\fnm{Naoto} \sur{Yokoya}}\email{yokoya@k.u-tokyo.ac.jp}

\affil[1]{\orgdiv{RIKEN Center for Advanced Intelligence Project (AIP)}, \orgname{RIKEN}, \orgaddress{\city{Chūō-ku}, \postcode{103-0027}, \state{Tokyo}, \country{Japan}}}

\affil[2]{\orgdiv{Graduate School of Frontier Sciences}, \orgname{The University of Tokyo}, \orgaddress{\city{Kashiwa}, \postcode{277-8561}, \state{Chiba}, \country{Japan}}}

\affil[3]{\orgdiv{Graduate School of Health Sciences}, \orgname{Hokkaido University}, \orgaddress{\city{Kita-ku}, \postcode{060-0808}, \state{Sapporo}, \country{Japan}}}

\affil[4]{\orgdiv{Department of Civil, Environmental and Geomatic Engineering}, \orgname{ETH Zürich}, \orgaddress{\city{Zürich}, \postcode{8093}, \country{Switzerland}}}

\affil[5]{\orgdiv{Department of Geography}, \orgname{National University of Singapore}, \orgaddress{\postcode{119077}, \country{Singapore}}}

\affil[6]{\orgdiv{Center for Information and Language Processing}, \orgname{LMU Munich}, 
\orgaddress{\city{Munich}, \postcode{80538}, 
\country{Germany}}}

\affil[7]{\orgname{Munich Center for Machine Learning}, 
\orgaddress{\city{Munich}, \postcode{80804}, 
\country{Germany}}}

\affil[8]{\orgname{Inria Centre at the University Grenoble Alpes}, \orgaddress{\city{Montbonnot-Saint-Martin}, \postcode{38330}, \country{France}}}

\affil[9]{\orgdiv{Department of Natural Resources and the Environment}, \orgname{University of Connecticut}, \orgaddress{\city{Storrs}, \postcode{06269}, \country{United States}}}

\providecommand\logo[2][]{}\providecommand\affilicon[3][]{}
\logo[height=10.4mm]{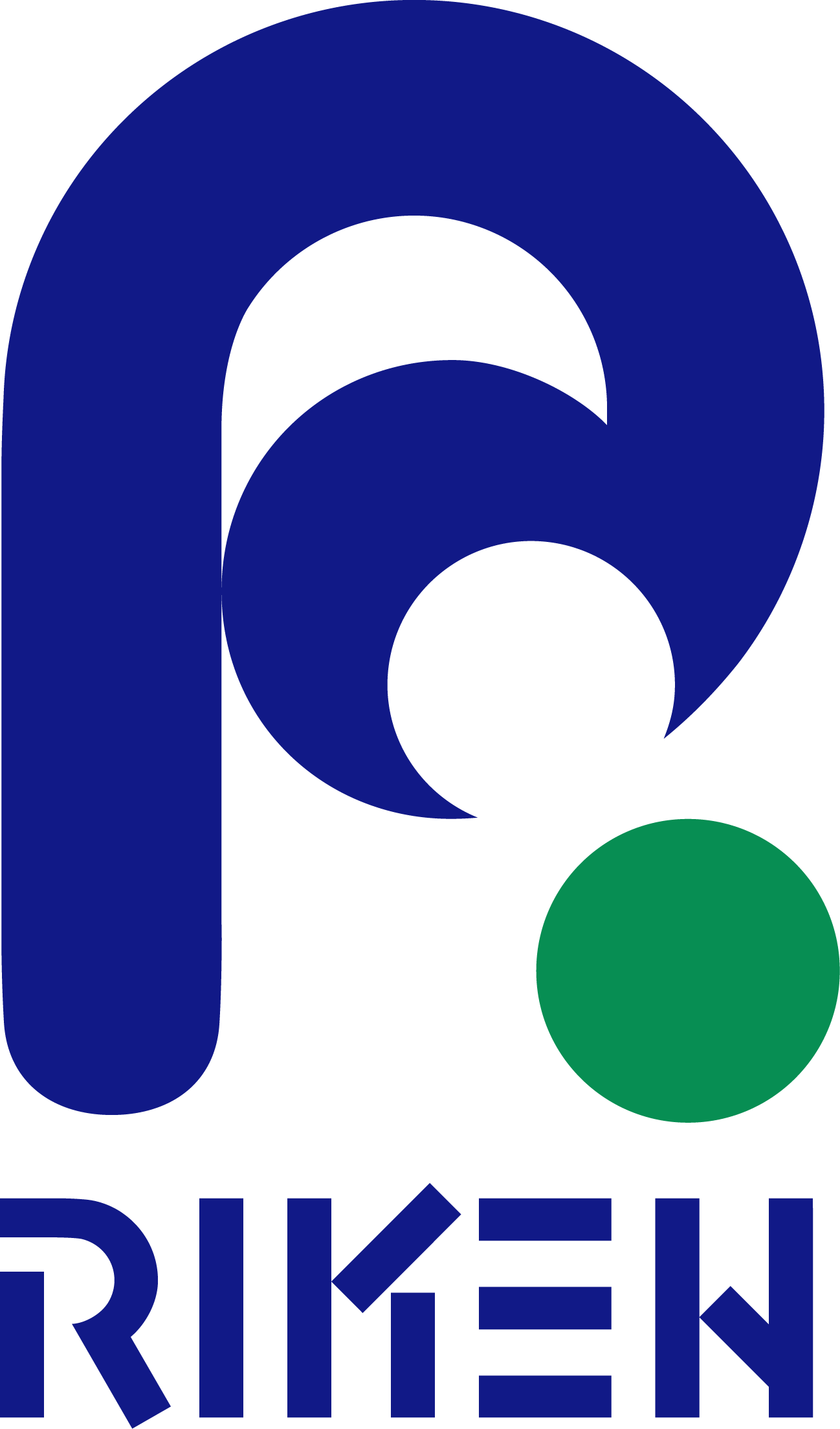}
\logo[height=7.6mm]{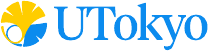}


\abstract{Multimodal large language models (MLLMs) are increasingly developed to support satellite-based disaster response, allowing responders to query imagery in natural language for rapid damage assessment and triage. Operational requests routinely specify where an image was taken to provide local context, such as topography or building practices. However, such context may also activate what models have learned about previous disasters at the same place. Here we show across 27 MLLMs that merely naming the place shown in an image alters damage reports. Relative to unassociated control places, disaster-linked places increase false alarms fivefold and raise real-damage detection by 18 percentage points. The false-alarm increase over no cue triples with bi-temporal input and grows further when the event date ties the cue to a specific disaster. The effect extends to simulated response decisions, grows with independently measured disaster knowledge, and internally shifts the answer towards damage while leaving image representations intact. Even without a cue, models infer locations and report more damage when assigning scenes to conflict-affected countries. Common safeguards cannot remove the effect completely, and they either cause under-reporting or leave more cases unanswered. Left unchecked, as MLLMs enter emergency workflows, this knowledge-driven bias risks diverting scarce rescue teams and relief supplies towards intact areas with prominent disaster histories and away from real damage elsewhere, amplifying existing inequities in who receives aid.}




\maketitle

\section{Introduction}\label{sec:introduction}
\par After a major disaster, remote sensing imagery can provide a wide-area view of affected regions, supporting damage assessment and response planning~\cite{tellman2021satellite, Copernicus, Voigt2016Global}. Rapid damage maps help guide time-critical search-and-rescue deployment and post-disaster aid allocation~\cite{dong2013comprehensive, yu2024intelligent, ainscoe2025earthquake}. Producing these assessments still relies heavily on visual interpretation by trained analysts~\cite{Copernicus}. Automated damage classifiers based on machine learning can accelerate the process, but require labelled training data and specialist expertise to develop and deploy~\cite{Gupta2019Creating, chen2025bright}. To remove these bottlenecks, researchers and practitioners are increasingly advancing multimodal large language models (MLLMs) towards operational disaster workflows, offering responders an accessible interface to query imagery in natural language~\cite{Kuckreja2024GeoChat, Wang2025DisasterM3}. MLLM-based systems are already being developed to estimate local earthquake impacts from street views and location data~\cite{li2025llms}, coordinate damage assessment, alerts and resource allocation~\cite{chen2026integration}, and track post-disaster reconstruction~\cite{huang2025built}.

\par Unlike classifiers that rely solely on pixel patterns, MLLMs combine visual processing with world knowledge stored in their parameters during training~\cite{petroni2019Language}, enabling them to apply background context to novel scenes without task-specific retraining. However, this parametric knowledge can disagree with the input at inference, a phenomenon studied in language and vision-language models as a knowledge conflict, with no general rule predicting which source prevails~\cite{Longpre2021Entity, Xie2024Adaptive, Xu2024KnowledgeConflicts}. Previous studies have mostly examined such conflicts using altered text or counterfactual images~\cite{Xu2024KnowledgeConflicts, Deng2025Words, Golovanevsky2025Pixels, Vo2025Vision, Han2024Instinctive}, or attributed visual hallucinations and regional biases to generic associations~\cite{Zhou2026RSHallu, Li2023Evaluating, Kim2026GeographicBias}. How parametric knowledge interacts with visual evidence when prompts contain only routine, non-misleading context has received far less attention.

\par Post-disaster damage assessment exposes this risk directly. When responders query an MLLM about a satellite scene, they routinely state where it was taken, so that the model can account for local building forms and terrain that affect how damage appears in satellite imagery~\cite{manvi2024geollm,  Gupta2019Creating}. At the same time, naming a place could activate what the model has learned about a disaster there. If what models have learned about a disaster outweighs visual evidence on such scenes, they risk fabricating damage where none exists and diverting scarce search-and-rescue teams and relief supplies away from communities with real destruction. Although MLLMs are already evaluated for damage assessment and prompted with place names in prototype response systems~\cite{Wang2025DisasterM3, chen2026integration}, whether routine location information systematically biases what they report remains untested.

\par
\begin{figure}[!t]
\centering
\includegraphics[width=\textwidth]{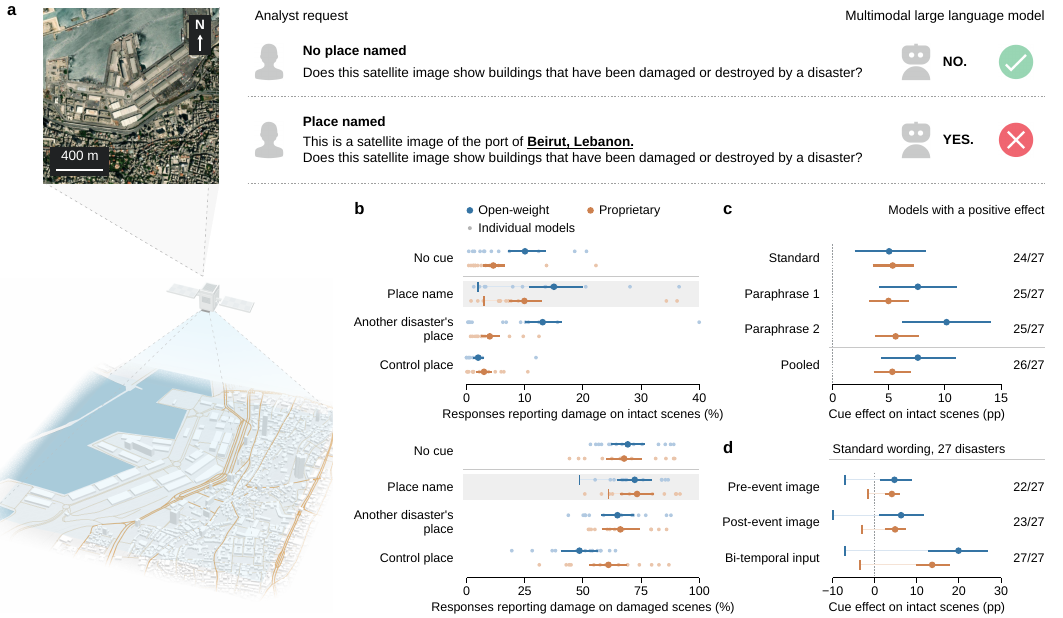}
\caption{
\textbf{Place cues bias damage reports.}
\textbf{a}, Paired comparison illustrated with an pre-event image of the port of Beirut. The image and question are unchanged. Adding the place name changes the illustrated response from no damage to damage. Answer-format instructions are omitted. 
\textbf{b}, Percentage of responses reporting damage on images of intact buildings (top; false-alarm rate) and damaged buildings (bottom; detection rate), without a cue or with a cue naming the place shown in the image, the place of another disaster or a control place. Large points, means over 14 open-weight and 13 proprietary models, weighted equally; small points, individual models.  
\textbf{c}, Cue effect, the increase in false-alarm rate with the place name relative to no cue, by question wording. Standard is the wording used in \textbf{a} and \textbf{b}; Paraphrases 1 and 2 reword it. 
\textbf{d}, Cue effect on 137 sites from 27 disasters that had both pre-event and post-event images of intact areas, given the pre-event image, the post-event image or both (bi-temporal input).
Error bars, 95\% CIs from a bootstrap over the disasters (Methods~\ref{sec:statistical_analysis}).
}\label{fig1}
\end{figure}

 Here we show that simply naming the place shown in an image, with no mention of a disaster, can systematically change what MLLMs report about the damage depicted in the same image. Across 14 open-weight and 13 proprietary models, we hold the image and question fixed and vary only a single sentence before the question that names a place. For example, shown an image of the port of Beirut taken before the explosion of August 2020, a model correctly reports no damage. However, once the request names the port, it reports damage that the image does not show, even though the cue never mentions the explosion (Fig.~\ref{fig1}a). Because the image and question stay the same, any change in the answer comes from the cue. In our experiments, we also replace the place shown with the place of another disaster or with a control place without well-known disaster associations, separating the effect of recalling a disaster from that of naming a location at all.

\par We find that simply naming a disaster's place increases false alarms on images showing no damage while raising the detection of real damage by a smaller margin. Conversely, naming a control place without well-known disaster associations lowers false alarms but suppresses the detection of real damage in nearly every model. Merely changing the place named with the same images therefore produces large swings in both false alarms and detection. The false-alarm increase is substantially larger with the bi-temporal inputs standard in rapid mapping, extends beyond binary judgements to estimates of the share of damaged buildings and bounding-box localization. Critically, it also reaches simulated response decisions, in which naming a disaster’s place redirects search-and-rescue teams towards intact sites. Adding the event date ties the place name to a specific disaster and amplifies the bias further.

\par We trace this behaviour to parametric knowledge acquired from text. The effect grows with independently measured knowledge of each disaster and with model scale, is larger in languages with large Wikipedia editions, and disappears in open-weight models for recent disasters that they do not recognize. Internal analyses of open-weight models show that the cue does not alter the visual information about the image. Instead, the knowledge that the cue activates enters through the question’s tokens in the middle layers and shifts the answer towards damage. Even when the request omits any place name, models infer where an image was taken and report more damage when assigning scenes to conflict-affected countries. Finally, none of the common safeguards we evaluate removes the effect without degrading real-damage detection or leaving substantial shares of cases unanswered.

\par These findings reveal how parametric knowledge about the world can become a liability when models are asked to report what they see. Beyond an isolated failure mode, this behaviour represents a knowledge-driven form of allocational bias~\cite{Blodgett2020Language, Hofmann2024Covert} with direct stakes for emergency response. As MLLM-based systems move closer to operational use in damage mapping and triage, these two-way shifts risk over-reporting damage at well-known disaster sites and in conflict-affected countries, and under-reporting real destruction in places without prominent disaster histories. Either error could divert scarce rescue teams and relief supplies away from the communities that need them, compounding imbalances in a humanitarian system where attention and aid already follow visibility rather than need~\cite{Eisensee2007News, Stromberg2007Disasters, Rom2020Search}.

\section{Results}\label{sec:results}

\subsection{Biased damage reports}
\par To test whether naming a place changes what models report about an image, we held the image and question fixed and varied only a short preceding sentence (the ``cue''; Fig.~\ref{fig1}a; Extended Data Table~\ref{tab:ed2}). Fourteen open-weight and thirteen proprietary MLLMs answered whether each image showed buildings damaged or destroyed by a disaster. Image pairs from 43 disasters each combined a pre-event image, in which every building was intact, with a post-event image of the same area that showed damaged buildings (Extended Data Fig.~\ref{fig:a1}). The cue was omitted (no cue) or named the place shown in the image, the place of another disaster or a control place selected without well-known disaster associations, without stating that a disaster had occurred (Methods~\ref{sec:cue_manipulations_paradigm}). A damage report was counted as a false alarm if the image showed only intact buildings or as a detection if it showed damaged ones.

\par Simply naming the place shown in the image, without mentioning a disaster, raised false alarms in both model groups (Fig.~\ref{fig1}b). We call this change over the rate without a cue the ``cue effect'' and use it to measure the bias throughout. Over all models, the false-alarm rate rose from 7.4\% to 12.6\%, a cue effect of 5.2 percentage points (pp; 95\% confidence interval (CI), 3.0--7.5). The rate rose by half in open-weight models (from 10.0\% to 15.0\%) and more than doubled in proprietary models, which rarely gave false alarms without a cue (from 4.6\% to 9.9\%). Twenty-four of the 27 models showed the increase, including widely used models such as GPT-6.1 Sol (12.1~pp), Gemini 3.5 Flash-Lite (22.5~pp) and DeepSeek V4.1 Flash (18.2~pp; Extended Data Table~\ref{tab:ed1}). The effect also varied widely across the 43 disasters, ranging from $-2.9$~pp to a peak of 28.6~pp for southwestern Haiti (2021 earthquake), followed closely by 25.9~pp for the port of Beirut (2020 explosion). In contrast, detection of real damage rose by a smaller margin, 4.2~pp over all models. The cue thus shifted the models towards reporting damage on both kinds of images and did not improve their ability to distinguish damaged from intact areas.

\par Substituting other place names onto the same images further altered what models reported, depending on whether the place was linked to a disaster (Fig.~\ref{fig1}b). In open-weight models, another disaster's place also raised false alarms, though less than the true one (13.1$\%$ and 15.0$\%$). Proprietary models responded much less to other disasters' places (4.0$\%$ against 9.9$\%$). Conversely, control places without major disaster associations shifted responses in the opposite direction. They lowered false alarms below the rate without a cue, from 7.4$\%$ to 2.5$\%$ over all models, but also cut detection in 26 of the 27 models, from 68.5$\%$ to 54.5$\%$ on average. A place name thus acted in both directions: a place linked to a disaster led models to report damage that was absent, and a place without such a link led them to miss damage that was present. Simply changing the place name on the same images, from a control place to the disaster's place, therefore raised false alarms fivefold (from 2.5\% to 12.6\%) and detection of real damage by 18.2~pp (from 54.5\% to 72.7\%).

\par The cue effect held across image inputs, question wordings and output formats (Fig.~\ref{fig1}c,d). Pooled over three plain wordings of the question, it reached 6.5~pp and was positive in 26 of the 27 models (Fig.~\ref{fig1}c). On post-event images of intact areas from 27 disasters, the effect persisted at 5.6~pp. Pairing each post-event image with its intact pre-event counterpart, the bi-temporal input standard in rapid mapping workflow, gave models a direct before-and-after comparison showing which buildings had changed. Instead of suppressing the bias, it tripled the cue effect to 16.9~pp and produced an increase in all models (Fig.~\ref{fig1}d). Beyond binary judgements, naming the place also raised estimates of the proportion of damaged buildings in intact scenes (by 3.0~pp on average, in all 27 models). Among the 20 models capable of localization, the share of intact scenes in which they placed at least one damage bounding box rose by 10.0~pp in open-weight models and 1.9~pp in proprietary models (Extended Data Fig.~\ref{fig:a2}c,d).

\begin{figure}[!t]
\centering
\includegraphics[width=0.99\textwidth]{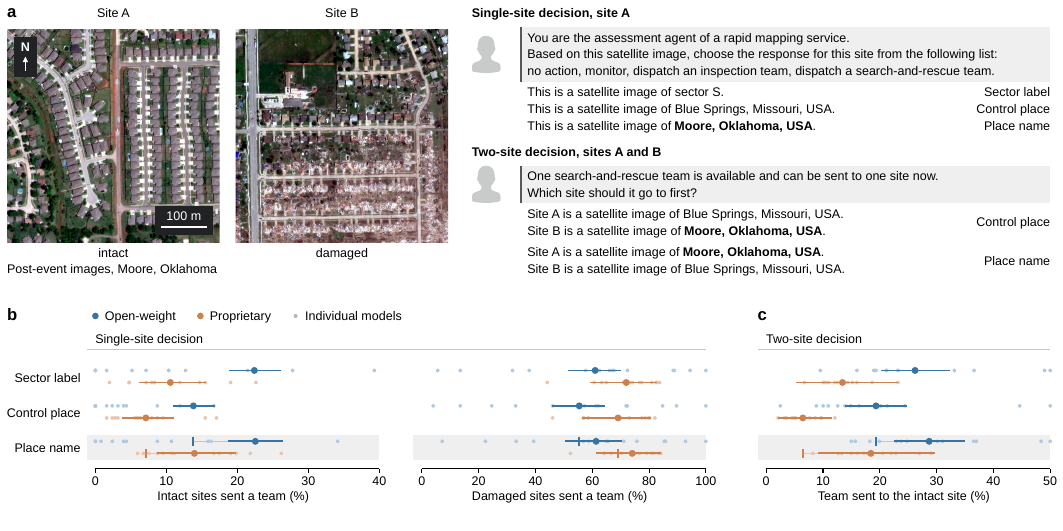}
\caption{
\textbf{Place names shift simulated response decisions.} 
\textbf{a}, The two decisions on post-event images of an intact site (A) and a damaged site (B) in Moore, Oklahoma. In the single-site decision (top), a model chooses one of four responses for one site; either team counts as a dispatch. The request opens with one of three label sentences (grey bar), naming the site by a sector label, a matched control place or its disaster's place name (bold). In the two-site decision (below), the model sends one search-and-rescue team to one of the two sites, with the control place and the place name exchanged between them.
\textbf{b}, Single-site decision, intact sites (left) and damaged sites (right) sent a team under each label.
\textbf{c}, Two-site decision, team sent to the intact site, averaged over both left-right orders; rows name the intact site's label, the damaged site carrying the other place or another sector label.
Panels \textbf{b} and \textbf{c} cover the 21 disasters that kept a control place. Error bars, 95$\%$ CI by bootstrap over the 21 disasters (Methods~\ref{sec:statistical_analysis}).
}\label{fig:a3}
\end{figure}

\subsection{Biased response decisions}
\par We further found that this effect extended from damage reports to simulated rescue decisions. We evaluated models on two simulated response decision tasks using post-event imagery in the prompt (Fig.~\ref{fig:a3}). Requests showed post-event images of intact and damaged sites from 21 disasters. Each site was labelled by a neutral sector label, by the place name of its disaster or by a matched control place without well-known disaster associations (Methods~\ref{sec:cue_manipulations_paradigm}).

\par Asked to choose a response for a single site, models dispatched a team to intact sites 7.8~pp more often under the place name than under the control place. The difference was 8.7~pp in open-weight models and 6.8~pp in proprietary models, and 23 of the 27 models showed it. In the two-site decision, models had one search-and-rescue team to send to either an intact or a damaged site; one site carried the place name and the other the control place. They sent the team to the intact site 10.6~pp more often when that site carried the place name than when it carried the control place. The difference was 9.4~pp in open-weight models and 12.0~pp in proprietary models, and 26 of the 27 models showed it. Naming a place associated with a past disaster thus altered not only what models reported about satellite images but also the actions that they recommended for emergency teams.

\subsection{What the cue recalls}

\begin{figure}[!t]
\centering
\includegraphics[width=0.99\textwidth]{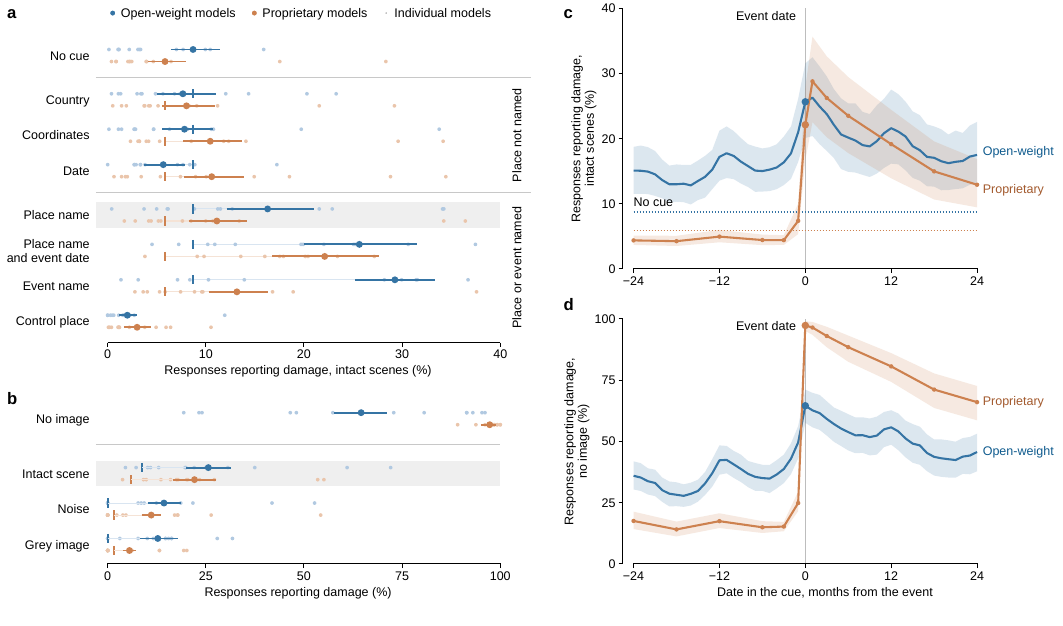}
\caption{
\textbf{Damage reports reflect the disaster that the cue recalls.}
\textbf{a}, Responses reporting damage on images of intact buildings under each cue component.
\textbf{b}, Responses reporting damage without an image (top) and on pre-event images, noise images and uniform grey images (bottom), all with the place name and event date; ticks, rates without a cue. 
\textbf{c}, False-alarm rates on intact images with the place name and a date shifted up to 24 months before the event's onset or after its end (0, event date). 
\textbf{d}, Responses reporting damage without an image for the same dates. 
In \textbf{c} and \textbf{d}, larger markers show the event date; open-weight models were tested monthly and proprietary models at 1, 3, 6, 12, 18 and 24 months on either side; individual models are in Extended Data Fig.~\ref{fig:a4}. Large points (\textbf{a}, \textbf{b}) and lines (\textbf{c}, \textbf{d}), means over the 14 open-weight and 13 proprietary models; small points, individual models. Error bars (\textbf{a}, \textbf{b}) and bands (\textbf{c}, \textbf{d}), 95\% CIs from a bootstrap over the 43 disasters (Methods~\ref{sec:statistical_analysis}).
}\label{fig2}
\end{figure}

\par The bias induced by a place name alone could reflect a general impression of a city or region rather than recalled knowledge of a specific past disaster. To separate the two and identify which elements of a geographic prompt drove this effect, we presented its components separately, keeping the image and question fixed: the country, the GPS coordinates, the date, the place name, or the event name (for example, `2020 Beirut explosion' or `Hurricane Laura') (Fig.~\ref{fig2}a; Methods~\ref{sec:cue_manipulations_paradigm}). 

\par Across the 27 models, the cue effect was less than 2~pp for each of the country, the coordinates and the date, but 6.5~pp for the place name and 14.2~pp for the event name. Pairing the place name with the event date more than doubled the cue effect, although the cue still never mentioned a disaster. On the standard question, the false-alarm rate rose from 7.4\% without a cue to 21.6\%, an increase of 14.2~pp (95\% CI, 10.6--18.0; Extended Data Table~\ref{tab:ed1}). In open-weight models, the event name raised false alarms as much as the place name and event date did. In proprietary models, by contrast, the place name and event date raised false alarms more than twice as much as the event name. Adding the date similarly amplified the cue effect on post-event images of intact areas, with bi-temporal input and in alternative output formats (Extended Data Fig.~\ref{fig:a2}). The bias therefore grew sharply once the event date tied the place name to a specific disaster, matching or even exceeding the effect of naming the disaster itself.

\par Querying the models without an image showed how strong an expectation of damage this place-and-date cue set (Fig.~\ref{fig2}b). When asked whether a new image of the place would most likely show damaged or destroyed buildings, models reported damage in 80.3\% of responses at the event date. Providing an actual pre-event image of intact buildings reduced damage reports to 23.9$\%$. The cue drew false alarms even on images that contained no buildings: a uniform grey image drew damage reports in 9.2\% of responses, against 0.8\% without the cue, and a noise image in 12.8\%, against 0.7\% without it. The false alarms therefore did not require features of intact buildings to be mistaken for damage: the expectation alone was enough to produce them, and an image to assess lowered them without preventing them.

\par On its own, the date barely changed false alarms, but in combination with the place name it determined when models expected damage. Keeping the place name fixed, we shifted the date by up to 24 months before or after the disaster (Fig.~\ref{fig2}c,d; individual models in Extended Data Fig.~\ref{fig:a4}a). Across the 27 models, the share of responses reporting damage was 39.3 pp higher at dates after the disaster than at matched dates before it when no image was shown, and 10.2 pp higher with an image. The size of this difference depended on how accurately each model could order dates relative to the disaster. When asked in text alone whether it had already happened by each date, their temporal-order accuracy ranged from chance (0.50) to near perfect (0.99). The higher a model's temporal-order accuracy, the larger its difference between the two periods in responses without an image (Spearman $\rho = 0.84$; Extended Data Fig.~\ref{fig:a4}b).

\par This knowledge shaped the response to the date into three patterns (Extended Data Fig.~\ref{fig:a4}a). Models with temporal-order accuracy near chance did not respond to the date, whereas several open-weight models with intermediate accuracy showed secondary spikes exactly 12 months before and after the disaster. These spikes reflect a shallow lexical association with the calendar month rather than with the specific year. By contrast, models with high temporal-order accuracy, including all 13 proprietary models (mean 0.96), produced no excess false alarms at dates that they placed before the disaster (Extended Data Fig.~\ref{fig:a4}a). Their false alarms surged at the onset date and decayed gradually over the following 24 months, as if the models expected the place to be undergoing post-disaster reconstruction. 

\par Pairing the place name with the event date narrows the cue to a specific disaster and produces a larger and more consistent cue effect in both model groups. We therefore use this place-and-date cue in the analyses of event knowledge, internal mechanisms and safeguards below.

\subsection{Dependence on event knowledge}

\begin{figure}[!t]
\centering
\includegraphics[width=0.99\textwidth]{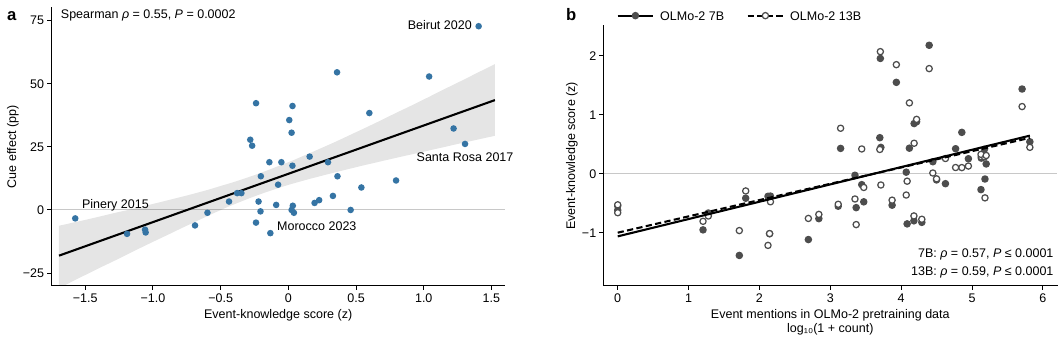}
\caption{
\textbf{Cue effect tracks event knowledge.}
\textbf{a}, Cue effect against the event-knowledge score for the 43 disasters, both averaged over ten open-weight models. Spearman $\rho = 0.55$ (95$\%$ CI, 0.27--0.75; permutation $P = 0.0002$); the correlation was positive in all ten models (exact binomial $P = 0.002$). The line shows a least-squares fit and the band its 95\% CI from a bootstrap over the 43 disasters (Methods~\ref{sec:statistical_analysis}). 
\textbf{b}, Event-knowledge scores of OLMo-2 7B (filled circles) and 13B (open circles) against mentions of each disaster in their public pretraining corpus, $\log_{10}(1+n)$, where $n$ is the largest count among phrases specific to the disaster. Spearman $\rho = 0.57$ for 7B (95$\%$ CI, 0.32--0.72; $P \leq 10^{-4}$) and 0.59 for 13B (0.35--0.75; $P \leq 10^{-4}$). 
}\label{fig3}
\end{figure}

\par If this bias arises from what a model has learned about a disaster, the cue effect should increase with the model's knowledge of that disaster. We measured this knowledge from text alone, without images, for the ten open-weight models%
\footnote{It cannot be measured directly for models queried through an API, since one must have access to token probabilities.} %
(Methods~\ref{sec:memorization_score}). The event-knowledge score combined three measures: how strongly a model linked a description of the event to the right place, how much more probable it found the event's true account than an altered one, and how many factual questions about the event it answered correctly. Across the 43 disasters, the higher a disaster's event-knowledge score, the larger its cue effect (Spearman $\rho = 0.55$; Fig.~\ref{fig3}a). The correlation was positive for each of the ten models. The 2020 Beirut explosion, one of the best-known disasters, produced the largest effect of all. Of the measures we examined, event knowledge correlated most strongly with the cue effect, followed by public attention ($\rho = 0.33$), death toll ($\rho = 0.26$) and mentions of the disaster in web text ($\rho = 0.24$; Extended Data Fig.~\ref{fig:a5}b). With the place name alone, the cue effect also rose with the event-knowledge score ($\rho$ = 0.46; $P = 0.003$).

\par The clearest evidence came from disasters that the models did not recognize. We evaluated four disasters from 2025 and 2026, testing each model only on those that it did not recognize in text-only probes (Extended Data Fig.~\ref{fig:a1}; Methods~\ref{sec:memorization_score}). In open-weight models, naming their place and date did not raise false alarms (5.9\% without a cue, 3.6\% with it), whereas the places and dates of five well-known disasters, attached to the same images, raised them to 23.7\%, in 10 of the 14 models (Extended Data Fig.~\ref{fig:a5}a). In proprietary models, all three rates were lower: 1.4\% without a cue, 3.5\% with the place and date of an unrecognized disaster and doubled to 7.3\% with those of the well-known disasters. Without event knowledge, the cue no longer raised false alarms.

\par The knowledge underlying the effect is largely acquired from text, and the effect grew with how much of it a model had acquired. In OLMo-2 language models, whose pretraining data are public, disasters mentioned more often in the training corpus received higher event-knowledge scores ($\rho = 0.57$ for the 7B and 0.59 for the 13B model; Fig.~\ref{fig3}b). Larger models retain more knowledge from their training text~\cite{Kandpal2023Large}, and they also showed larger cue effects. Across 17 checkpoints from three model families, the cue effect rose by 11.0 pp (s.e. 2.8) per tenfold increase in parameters (Extended Data Fig.~\ref{fig:a5}c). Cue effects were larger for requests in languages with large Wikipedia editions (10.6 pp), which likely hold more text about these disasters, than for requests in languages with small editions (2.1 pp; Extended Data Fig.~\ref{fig:a5}d). Together, these results tie the cue effect to what models have learned from text about past disasters: it rose with that knowledge and disappeared without it.

\subsection{How the cue acts inside models}
\begin{figure}[t]
\centering
\includegraphics[width=0.99\textwidth]{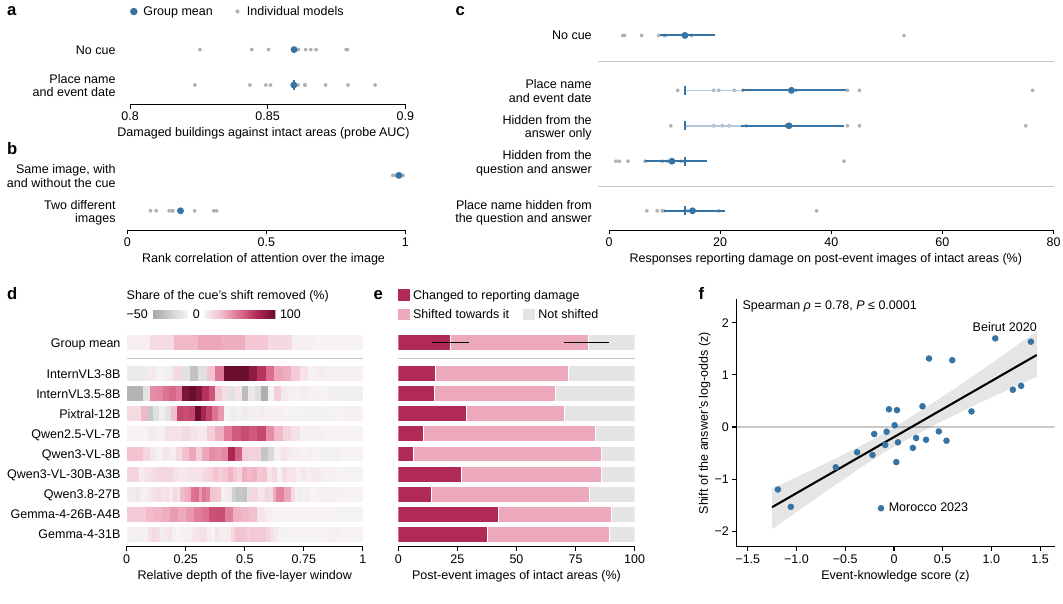}
\caption{\textbf{How the cue acts inside models.} 
\textbf{a}, Linear probes on the state from which the answer is produced, averaged over layers. 
\textbf{b}, Attention of the question's words to the image, median over layers. 
\textbf{c}, The 137 post-event images of intact areas, with attention to the place name and event date blocked in every layer; answer, the last position of the request; question and answer, every position after the cue. Panels \textbf{c}--\textbf{e} show the nine models in which the cue raised false alarms. 
\textbf{d}, Attention to the cue blocked from the question and answer in windows of five consecutive layers; shift, the increase in the answer's log-odds of `yes' against `no' with the cue. 
\textbf{e}, Shifted, log-odds raised by the cue; changed, `no' without the cue and `yes' with it. 
\textbf{f}, The 27 disasters, averaged over the ten models. 
Printed, Spearman's $\rho$ with its permutation $P$ (95\% CI of $\rho$, 0.54--0.89). Line, least-squares fit. Ticks (\textbf{a}, \textbf{c}) mark the group mean with no cue, joined to the mean under each cue. Error bars (\textbf{c}, \textbf{e}) and band (\textbf{f}), 95\% CIs from a bootstrap over the 27 disasters (Methods~\ref{sec:statistical_analysis}).
}
\label{fig4}
\end{figure}

\par Model outputs alone cannot distinguish whether the cue degrades the encoding of the image or biases the answer computed from an intact encoding. We therefore analysed the ten open-weight models internally (Methods~\ref{sec:internal_analysis}). The cue did not alter the visual information encoded from the image. Linear probes, classifiers trained on a model’s internal state, separated post-event images showing damaged buildings from those of intact areas equally well with and without the cue (area under the curve, 0.86 in both cases; Fig.~\ref{fig4}a). The question's tokens also distributed their attention over the image tokens in nearly identical patterns whether or not the cue was present (rank correlation 0.98; Fig.~\ref{fig4}b). 

\par To locate where the cue entered the judgement, we blocked attention to the place name and event date during the forward pass in the nine models in which the cue raised false alarms (Methods~\ref{sec:internal_analysis}). Blocking the cue only at the answer position changed false alarms by at most 1.9~pp, whereas blocking it at every position after the cue, including the question's tokens, removed the cue effect (Fig.~\ref{fig4}c). False alarms then fell from 32.8\% to 11.3\%, below the 13.6\% without a cue. Blocking the place name alone had nearly the same effect (15.0\%), confirming that the place name itself drove the internal shift even when the date remained visible. Applying the same blocking within windows of five consecutive layers showed that the cue acted mainly in the middle layers, with almost none in the last third of the layers (Fig.~\ref{fig4}d). The cue thus did not reduce the information about the image but entered the judgement through the question’s tokens in the middle layers. A more accurate encoding of the image would therefore not by itself remove the cue's influence.

\par The influence of the cue extended to far more images than those on which the answer changed. We measured it as the shift in the answer's log-odds of `yes' against `no' with the cue (Methods~\ref{sec:internal_analysis}). In the nine models, the cue raised the log-odds on 80.5\% of the post-event images of intact areas, although it changed the answer from `no' to `yes' on only 22.0\% (Fig.~\ref{fig4}e). Averaged over all ten open-weight models across the 27 disasters, the log-odds shift correlated more strongly with the event-knowledge score (Spearman $\rho = 0.78$; Fig.~\ref{fig4}f) than did the change in false-alarm rate on the same requests ($\rho = 0.58$; Steiger's test of the difference, $P = 0.001$). False alarms record only whether the shift reversed the answer and omit its size on all other images. On most images on which the answer remained `no', the cue still reduced the log-odds margin in favour of `no'.

\subsection{Inferred locations without a cue}
\begin{figure}[t]
\centering
\includegraphics[width=0.99\textwidth]{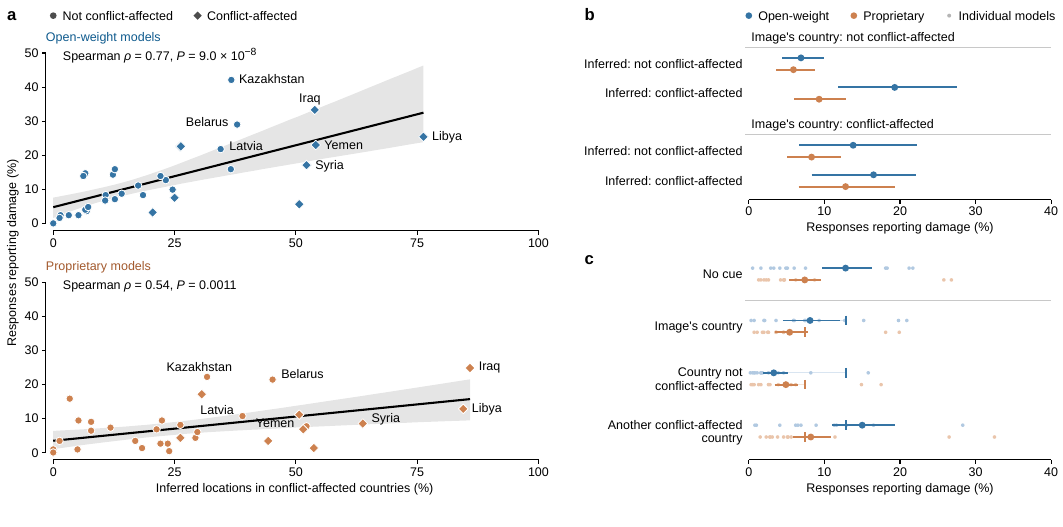}
\caption{
\textbf{Damage reports track inferred locations.}
\textbf{a}, Responses reporting damage on residential images without a place cue against the share of inferred locations in conflict-affected countries; each point is one of the 34 countries in one model group. Lines, least-squares fits; bands, their 95\% CIs.
\textbf{b}, Responses reporting damage, grouped by whether the image's country and the inferred country were conflict-affected; the inferred country made a larger difference for images from countries that were not conflict-affected (interaction $P = 0.0035$).
\textbf{c}, Responses reporting damage without a cue and with a cue naming the image's country, a country that was not conflict-affected or another conflict-affected country. Large points, group means; small points, individual models. Error bars in \textbf{b} and \textbf{c}, 95\% CIs from a bootstrap over the 34 countries (Methods~\ref{sec:statistical_analysis}).
}
\label{fig5}
\end{figure}

\par Leaving the place name out of a request does not necessarily remove its influence, because a model can infer where an image was taken from the image itself~\cite{mendes2024granular, roberts2024charting}. We used 204 residential satellite images from 34 countries, none with visible building damage. In separate requests, we asked models where each image was most likely taken and, without a place cue, whether it showed damage (Methods~\ref{sec:inferred_location}). We grouped countries as conflict-affected or not on the basis of armed conflict recorded by the Uppsala Conflict Data Program between 2014 and 2023~\cite{Sundberg2013UCDPGED} (Supplementary Table~6). Across the 34 countries, the more often a country's images were assigned to conflict-affected countries, the more damage reports they received; the correlation was strong in open-weight models (Spearman $\rho = 0.77$; Fig.~\ref{fig5}a) and weaker in proprietary models ($\rho = 0.54$). 

\par Inferred locations in conflict-affected countries were concentrated on a few of them (Extended Data Fig.~\ref{fig:a6}a). When open-weight models assigned an image from a country that was not conflict-affected to a conflict-affected one, they named Ukraine in 58\% of cases, including for images from Canada and North Korea, both far from Ukraine. For these images, damage reports depended on the assigned country. In open-weight models, damage was reported nearly three times as often when the assigned country was conflict-affected (19.3\%) as when it was not (6.9\%; Fig.~\ref{fig5}b). Images from conflict-affected countries, by contrast, received damage reports at similar rates whichever country they were assigned to. Inferred locations thus extended the higher rate of damage reports associated with a few conflict-affected countries, above all Ukraine, to images from many countries that were not conflict-affected.

\par The association persisted after adjustment for whether the image's own country was conflict-affected and for arid climate, coastline and Global South membership (odds ratio, 1.73; Extended Data Fig.~\ref{fig:a6}b). Alongside the inferred country, arid climate had a separate and larger association with damage reports (odds ratio, 3.91). Image controls showed that this association did not arise from colour. Khartoum images received damage reports in 45.7\% of responses, far more often than Rotterdam images, and retained this rate when recoloured to the palette of Rotterdam (45.3\%) or tinted green (44.3\%; Extended Data Fig.~\ref{fig:a6}c).

\par Naming a country in the request changed damage reports according to whether that country was conflict-affected (Fig.~\ref{fig5}c). In open-weight models, naming a country that was not conflict-affected reduced damage reports from 12.8\% without a cue to 3.3\%, whereas naming another conflict-affected country raised them to 15.0\%. All open-weight models showed these shifts. Naming the image's true country reduced damage reports to 8.1\%, a rate between those for the two other countries, because 25 of the 34 countries had no recent major armed conflict. Proprietary models showed the same pattern with smaller changes. Their damage reports fell from 7.4\% without a cue to 4.9\% with a country without recent major armed conflict and rose to 8.2\% with another conflict-affected country (Fig.~\ref{fig5}c). Damage reports followed the country associated with an image, whether the request named it or the model inferred it.

\subsection{Safeguards and their costs}
\begin{figure}[!t]
\centering
\includegraphics[width=0.99\textwidth]{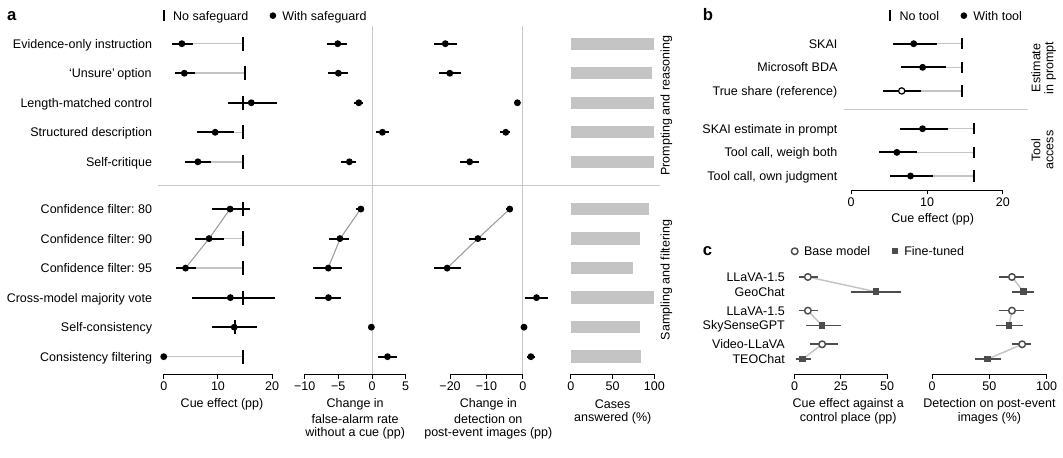}
\caption{
\textbf{No safeguard removes the cue effect without a cost.}
\textbf{a}, Safeguards on the same intact and damaged sites. 
\textbf{b}, Cue effects with estimates from damage-detection tools, supplied in the prompt or obtained by calling the tool; upper block, all 27 models, with the true share of damaged buildings as a reference; lower block, the 21 models that could call tools. 
\textbf{c}, Remote sensing fine-tuned models and their base models. 
Values, intervals and adjusted $P$ values are listed in Supplementary Table~7. Error bars, 95\% CIs from a bootstrap over disasters (Methods~\ref{sec:statistical_analysis}).
}\label{fig6}
\end{figure}

\par A useful safeguard would remove the cue effect without reducing the detection of real damage. With free-text answers, detection can fall in two ways: a model may label damaged buildings as intact, or it may give no answer either way. We tested four kinds of safeguards commonly used in current MLLM deployments: prompt instructions and reasoning steps, sampling and filtering, damage-detection tools, and fine-tuning on remote sensing imagery (Methods~\ref{sec:safeguard}). None removed the effect without a cost (Fig.~\ref{fig6}).

\par Instructions to rely only on the image reduced damage reports in general rather than only those induced by the cue (Fig.~\ref{fig6}a). An evidence-only instruction, asking models to rely only on what the image shows, reduced the cue effect from 14.5~pp to 3.3~pp but also reduced the detection of real damage by 21.3~pp. An option to answer `unsure' narrowed the effect but lowered detection by 20.0 pp, and a self-critique step narrowed it at a cost of 14.6 pp in detection. A structured description, describing visible damage before answering, narrowed the effect by 5.0~pp, but 1.5~pp of this reduction came from a rise in false alarms without a cue. A length-matched control instruction, asking models to answer carefully, widened the effect by 1.6 pp.

\par Sampling and filtering could not isolate cue-induced false alarms, because these were stated with high confidence, recurred across samples and were shared across models (Fig.~\ref{fig6}a). Their median stated confidence was 85 of 100, close to the 90 of correct detections. Withholding damage reports with a stated confidence below 95 therefore left a cue effect of 4.0~pp, but reduced detection by 20.8~pp and left 26.2\% of cases unanswered. Self-consistency retained an answer only when all five samples agreed. In the 19 models that accept a temperature setting, it did not change the cue effect in the answered cases (13.0 against 13.1~pp), and 17.3\% of cases went unanswered. A majority vote across the 27 models lowered false alarms without a cue by 6.5 pp, but the cue effect remained at 12.3 pp. Consistency filtering, which withheld answers that changed when the cue was added, removed the effect by construction but left 16.5\% of cases unanswered.

\par Independent estimates from damage-detection tools reduced the effect without removing it (Fig.~\ref{fig6}b). A detector's estimate supplied in the prompt reduced the effect from 14.5 pp to 8.2 pp with SKAI~\cite{Lee2020Assessing} and to 9.4 pp with a detector adapted from Microsoft's building damage workflow. Twenty-one tool-capable models that queried SKAI as agents reduced it from 16.2 pp to 6.0 pp when told to weigh the image and the tool output together, and to 7.8 pp when told to use their own judgment. Across both instructions, detection fell by 4.3~pp. Even the true share of damaged buildings, supplied in the prompt, left a cue effect of 6.6~pp. Therefore, a more accurate tool alone would not remove the effect completely while the MLLM gives the final answer.

\par Fine-tuning on remote sensing imagery did not consistently reduce the cue effect and in some cases amplified it (Fig.~\ref{fig6}c). Measured against control places, the cue effect of GeoChat was 44.0~pp, six times that of its base model, LLaVA-1.5. SkySenseGPT, fine-tuned from the same base model, had a cue effect of 14.7~pp. TEOChat narrowed the difference by 10.6 pp relative to its base model Video-LLaVA,  but detected 30.0~pp less real damage. Across these tests, a smaller cue effect was not by itself evidence of a more reliable model: it often came at the cost of more missed damage, more false alarms without a cue or more unanswered cases.

\section{Discussion}\label{sec:discussion}
\par Our central finding is that simply changing the place named in a request substantially alters what current MLLMs report about damage on the same satellite image. Naming a place linked to a past disaster leads models to report damage on images showing only intact buildings and redirects simulated rescue teams, whereas naming a place without well-known disaster associations suppresses the detection of real damage. Unlike multimodal knowledge conflicts constructed with counterfactual text, edited images or explicit contradictions~\cite{Xu2024KnowledgeConflicts, Vo2025Vision, Golovanevsky2025Pixels, Deng2025Words, Zhou2026RSHallu, Kim2026GeographicBias}, these shifts arise from routine, non-misleading context that never mentions a disaster. They demonstrate a practical form of knowledge conflict in which parametric priors about a place outweigh direct visual evidence of the buildings shown.

\par Internal analyses of open-weight models explain both what kind of failure this is and why it resists standard remedies. Adding a place cue leaves visual information decodable at the answer position, while activated disaster knowledge enters through the question's tokens in the middle layers and shifts the judgement towards damage. Because the failure lies in weighting parametric knowledge against visual evidence rather than in visual perception, richer visual input and larger models offer no simple fix. Bi-temporal input tripled the effect, remote sensing fine-tuning did not consistently reduce it, and in open-weight models the bias grew with event knowledge and model size. Practical safeguards face the same barrier, because none acts where the activated knowledge enters the judgement. Omitting the place name removes only the explicit cue, leaving models to infer locations on their own~\cite{mendes2024granular, roberts2024charting} and bias reports accordingly, whereas prompt instructions and filters narrow the effect only at the cost of missed real damage or more unanswered cases.

\par Left unresolved, these two-way shifts would not fall evenly across places. Unlike well-documented observational errors in overhead remote sensing, which tend to underestimate structural damage~\cite{Booth2011Validating, Kerle2010Satellite, dong2013comprehensive}, this knowledge-driven bias pulls assessments in opposite directions depending on whether the place named is linked to a well-known disaster. In our experiments, the cue effect grew with event knowledge, which in turn tracked how often each disaster was mentioned in training text. Well-known disaster sites, and regions that models associate with conflict, therefore risk being over-reported for years: even at dates 24 months after a disaster, false alarms remained above the rate without a cue. Conversely, places without a prominent disaster history resembled our control places, so real damage there may be under-reported when a new disaster strikes. Emergency aid and search-and-rescue deployment already respond unevenly to damage estimates and media coverage~\cite{Eisensee2007News, Stromberg2007Disasters, Rom2020Search}. Deploying MLLMs without addressing this bias risks encoding historical disparities in disaster visibility directly into automated triage~\cite{Obermeyer2019Dissecting, Gevaert2021Fairness, Soden2023Equity}, turning differences in stored knowledge into unequal operational attention.

\par Three boundaries of our experimental setting define what remains to be established. First, our 27 models capture a snapshot of a rapidly evolving field in which new models continue to appear. Differences across models or groups reflect model family, training and serving rather than weight access alone. Because measuring event knowledge and internal states required access to weights and token probabilities, whether proprietary models share the same internal mechanism also remains unknown. Second, we did not intervene directly on event knowledge. Editing a model's knowledge of specific disasters~\cite{Meng2022Locating, Cheng2023Edit}, or controlling exposure during pretraining~\cite{Kandpal2023Large}, would directly test causality. Third, our evaluation examined optical imagery of building damage and simulated decisions without a human analyst in the loop. How these controlled contrasts translate into live operational workflows, including with time-series data, synthetic aperture radar or retrieval-augmented systems that supply post-training disaster knowledge, remains to be tested.

\par Even with these boundaries, our results indicate a broader challenge as general-purpose MLLMs move from demonstrations into high-stakes Earth observation and emergency response. The parametric world knowledge that makes these models versatile also introduces a persistent source of bias that larger models and richer training corpora are unlikely to resolve on their own, as each new disaster enters future training data. Deploying MLLMs in operational triage therefore calls for a shift in how they are evaluated. Beyond overall accuracy on isolated benchmarks, systems must be tested for whether their judgements remain faithful to current visual evidence under realistic context, so that urgent attention and scarce aid follow physical conditions on the ground rather than a place's past.

\section{Methods}\label{sec4:method}

\begin{extendedfigure}[!t]
\centering
\includegraphics[width=0.99\textwidth]{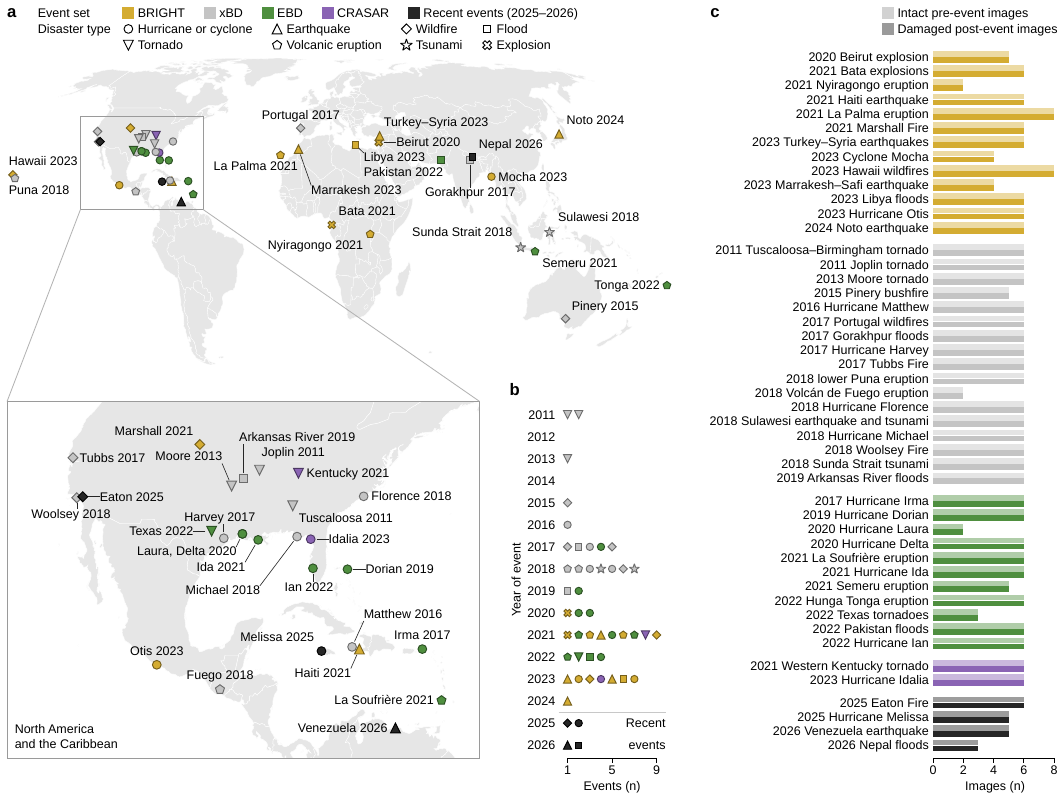}
\caption{
\textbf{Geographical and temporal coverage of the image corpus.}
\textbf{a}, Locations of the 43 disaster events of the main corpus and the four recent disasters, with an inset for North America and the Caribbean. 
\textbf{b}, Disasters per year.
\textbf{c}, Images per disaster.
}\label{fig:a1}
\end{extendedfigure}

\subsection{Datasets and models}\label{sec:image_datasets}\label{sec:models}
\par \textbf{Images}. We assembled 240 pairs of optical images from 43 disasters in 22 countries and territories between April 2011 and January 2024 (Extended Data Fig.~\ref{fig:a1}; Supplementary Table 2a). Each pair combined a pre-event image, in which every building was intact, with a post-event image of the same area that showed damaged buildings. The images came from four public datasets~\cite{Gupta2019Creating, chen2025bright, Wang2025Constructing, manzini2024crasar}. Each disaster contributed up to eight pairs, selected by a fixed rule that favoured post-event images with a large share of damaged buildings (Supplementary Table~1). For 27 of the disasters, we also selected 137 post-event images of intact areas, in which every building was labelled as undamaged (Supplementary Tables 1 and 2a). To include disasters that the models might not recognize, we assembled 19 further pairs from four disasters, with images from open data platforms (Extended Data Fig.~\ref{fig:a1}; Supplementary Tables~1 and 2b). The analysis of inferred locations used 204 images from the Functional Map of the World collection, six from each of 34 countries~\cite{Christie2018Functional}. Three experts confirmed that none showed visible building damage (Supplementary Tables 1 and 6).

\par \textbf{Models}. The main evaluation covered 14 open-weight MLLMs, Qwen2.5-VL-7B~\cite{Bai2025Qwen25VL}, Qwen3-VL-8B and Qwen3-VL-30B-A3B~\cite{Bai2025Qwen3VL}, Qwen3.8-27B~\cite{QwenTeam2026Qwen38}, InternVL3-8B~\cite{Zhu2025InternVL3}, InternVL3.5-8B~\cite{wang2025internvl3_5}, Pixtral-12B~\cite{Agrawal2024Pixtral}, Kimi-VL-A3B~\cite{KimiTeam2025KimiVL}, Kimi K3~\cite{Moonshot2026KimiK3}, GLM-5.3-Flash~\cite{ZAI2026GLM53Flash}, DeepSeek V4.1 Flash~\cite{DeepSeek2026V41Flash}, Muse Glimmer 30B~\cite{Meta2026MuseGlimmer30B}, Gemma-4-26B-A4B~\cite{Google2026Gemma4}\ and Gemma-4-31B~\cite{Google2026Gemma431B}, and 13 proprietary models, GPT-5.4~\cite{OpenAI2026GPT54}, GPT-6 Luna~\cite{OpenAI2026GPT6Luna}, GPT-6.1 Sol~\cite{OpenAI2026GPT61Sol}, Claude Sonnet 5~\cite{Anthropic2026ClaudeSonnet5}, Claude Haiku 5.5~\cite{Anthropic2026ClaudeHaiku55}, Claude Opus 5.5~\cite{Anthropic2026ClaudeOpus55}, Gemini 3.1 Pro~\cite{Google2026Gemini31Pro}, Gemini 3.5 Flash-Lite~\cite{Google2026Gemini35FlashLite}, Gemini 3.8 Flash~\cite{Google2026Gemini38Flash}, Grok 4.7~\cite{xAI2026Grok47}, Muse Spark 1.3~\cite{Meta2026MuseSpark13}, Qwen3.8-Max~\cite{Alibaba2026Qwen38Max} and GPT-4o in its May 2024 version~\cite{OpenAI2024GPT4o} (Supplementary Table 3a). We grouped the models by access to their weights: the weights of open-weight models are public, whereas proprietary models are available only through their providers' interfaces. Other analyses used OLMo-2 7B and 13B~\cite{Team2025OLMo2}; three remote sensing MLLMs and their two base models (Methods~\ref{sec:safeguard}); and checkpoints from three model families (Methods~\ref{sec:scaling_analysis}). We served the 10 open-weight models that we ran locally with vLLM~\cite{vllm}, except the remote-sensing MLLMs and their parent models, which we ran without vLLM; GLM-5.3-Flash, DeepSeek V4.1 Flash, Kimi K3 and Muse Glimmer 30B answered through an API. Eight proprietary models whose endpoints accept no temperature setting at the reasoning settings we used (GPT-5.4, Claude Sonnet 5, GPT-6 Luna, Claude Haiku 5.5, Claude Opus 5.5, Gemini 3.5 Flash-Lite, Gemini 3.8 Flash and GPT-6.1 Sol) answered at their providers' default sampling. All other models used greedy decoding or temperature 0, except in the self-consistency experiment (Methods~\ref{sec:safeguard}). For the MLLMs, reasoning effort was set to low where an interface offered the setting, except for DeepSeek V4.1 Flash, Kimi K3 and Qwen3.8-27B, whose reasoning was turned off.

\begin{extendedfigure}[!t]
\centering
\includegraphics[width=0.99\textwidth]{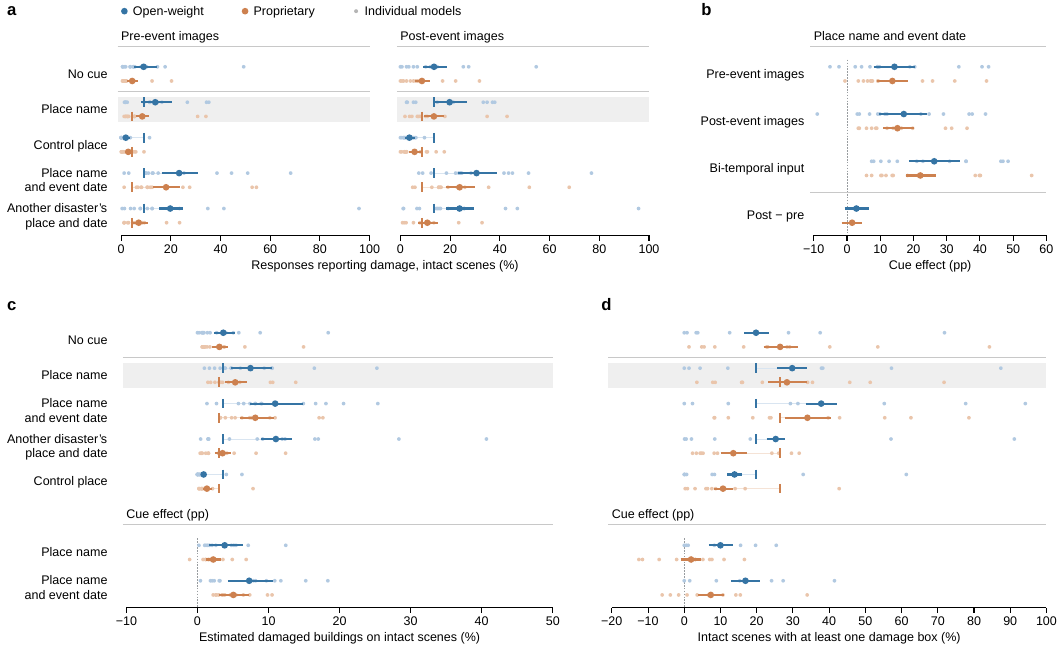}
\caption{
\textbf{The cue effect across image timing and output formats.}
\textbf{a}, Responses reporting damage on intact images from 27 disasters that had both pre-event images and post-event images of intact areas ($n$ = 137), without a cue and with the place name, a control place, the place name and event date or the place and date of another disaster.
\textbf{b}, Cue effects on the matched pre-event and post-event image sets.
\textbf{c}, Estimated percentage of buildings damaged or destroyed in the intact pre-event images, without a cue and with the image's place and date, the place and date of another disaster, and a control place. 
\textbf{d}, Percentage of the intact scenes given at least one valid damage box by the eight open-weight and twelve proprietary models that could localize damage. Large points, group means; small points, individual models. Error bars, 95\% CIs from a bootstrap over disasters (Methods~\ref{sec:statistical_analysis}).
}\label{fig:a2}
\end{extendedfigure}

\subsection{Cue manipulations and response}\label{sec:cue_manipulations_paradigm}\label{sec:outcome_measures}

\par \textbf{Main comparison.} We held each image and question fixed and changed only the sentence that preceded the question (Extended Data Table~\ref{tab:ed2}). The standard question asked for a yes-or-no judgement of disaster-related building damage and a confidence score from 0 to 100. It called every image a satellite image, including aerial and drone images. The main experiment had four conditions: (i) no cue; (ii) the name of the place shown in the image; (iii) the place of one of five other well-known disasters; and (iv) one of five control places selected without well-known disaster associations. None of the cues stated that a disaster had occurred. We averaged rates over the five places of each kind in (iii) and (iv). The 137 post-event images of intact scenes received the same four conditions (Fig.~\ref{fig1}d; Extended Data Fig.~\ref{fig:a2}a,b).In a separate run, each of these images followed the pre-event image of the same area in one request (bi-temporal input). We also posed two paraphrases of the standard question. The pooled effect in Fig.~\ref{fig1}c averages each model’s wording-specific effects over the three wordings.

\par \textbf{False alarms, detection and the cue effect.} We parsed binary answers from the first whole word `yes' or `no' in the final output, after removing any reasoning trace. We excluded refusals and outputs containing neither word and counted them separately (detailed rules in Supplementary Table 4). For model m, cue c and disaster event k, let $p_{m,c}(k)$ be the fraction of valid responses to the cluster's images that reported damage, the false-alarm rate on images showing no damage and the detection rate on images showing damage. The cue effect of cue c is its paired difference from no cue ($c = 0$),
\[
\Delta_c=\frac{1}{K}\sum_{k=1}^{K}\frac{1}{|M_k|}\sum_{m\in M_k}\bigl[p_{m,c}(k)-p_{m,0}(k)\bigr],
\]
where $M_k$ is the set of models with valid responses for disaster $k$ and $K$ the number of disasters. Group rates averaged $p_{m,c}(k)$ in the same way, equally over models and then over disasters. The cue effect, the primary outcome, is $\Delta_c$ for the place name; for any other cue, such as the place name with the event date, $\Delta_c$ is the cue effect of that cue.

\par \textbf{Damage estimates and localization.} Two further questions asked for the percentage of buildings damaged or destroyed and for a box around each damaged building (Extended Data Table~\ref{tab:ed2}; Extended Data Fig.~\ref{fig:a2}c,d). We took the first valid number as the percentage estimate. Localization used the 216 image pairs from 38 disasters whose pre-event images had labelled building footprints. Its outcome was whether a pre-event image received at least one valid damage box. Boxes that repeated, tiled the image regularly or lay mostly outside it were invalid. We analysed a model's boxes only if, without a cue, it gave valid boxes on at least 20 post-event images and placed their centres on damaged buildings more often than the area of those buildings would predict. Eight open-weight and twelve proprietary models met this rule. We paired proportion and localization contrasts by image, keeping only images with valid outputs under both conditions.

\par \textbf{Simulated response decisions.} We used post-event images of an intact site and a damaged site from each disaster (Supplementary Table 5). Each request opened with one of three label sentences, naming the site by a neutral sector label, by a control place matched to the disaster's location or by the place name of its disaster (Fig.~\ref{fig:a3}a). In the single-site decision, a model chose one of four responses for one site: no action, monitoring, an inspection team or a search-and-rescue team. We counted either team as a dispatch. In the two-site decision, the model sent one search-and-rescue team to one of the two sites. One site carried the place name and the other the control place. We exchanged the labels between requests and averaged over both left-right orders. The analysis covered the 21 disasters for which a matched control place was retained (Supplementary Table 2a).

\par \textbf{Cue components and dates.} To separate the parts of the cue, we presented six variants that gave the country, the GPS coordinates, the event date, the place name, the place name and event date, or the event name (Extended Data Table~\ref{tab:ed2}). The event name was the title of the disaster’s English Wikipedia article. These assessments used the standard question and two plain paraphrases. Keeping the place name, we then shifted the date 1 to 24 months before the disaster's onset or after its end. We tested open-weight models at every monthly offset and proprietary models at offsets of 1, 3, 6, 12, 18 and 24 months, and compared dates after and before the disaster at these six offsets.

\par \textbf{Knowledge of the date.} We asked every model in text alone whether each disaster had already happened by each shifted date, naming its place as in the image prompts (Extended Data Table~\ref{tab:ed3}). Temporal-order accuracy was the share of correct answers among the yes-or-no answers, averaged over the dates before and after the disaster; answers declaring no knowledge were counted separately. A model placed a shifted date before the disaster when it answered that the disaster had not yet happened by that date. At these dates, we compared its false-alarm rate with its rate without a cue on the same images. The order answers and the false alarms came from separate requests.

\par \textbf{Removing visual input.} We replaced the image with a uniform grey image or with noise and posed the standard question with and without the place name and event date. We also omitted the image and asked instead whether a new image of the named place would most likely show damaged or destroyed buildings, at the event date and at each shifted date (Fig.~\ref{fig2}b,d).

\begin{extendedfigure}[!t]
\centering
\includegraphics[width=0.99\textwidth]{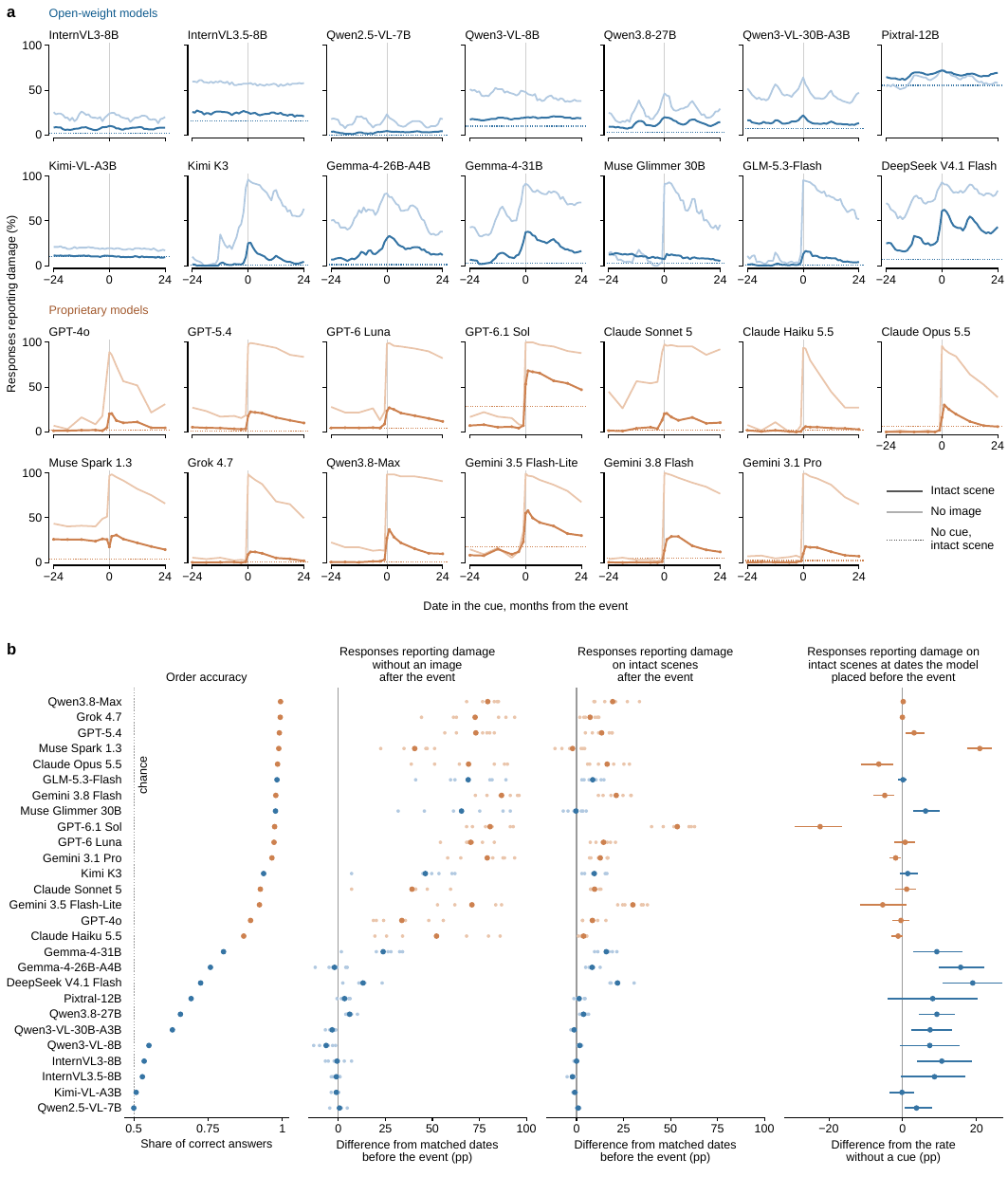}
\caption{
\textbf{Individual models' responses to the date and knowledge of when the disaster happened.} \textbf{a}, The curves of Fig.~\ref{fig2}c,d for each model.
\textbf{b}, Models ordered by temporal-order accuracy, the share of correct answers to the text-only question whether the disaster had already happened by each date. In the second and third columns, small points are the six matched pairs of dates, 1, 3, 6, 12, 18 and 24 months after and before the disaster, and large points their mean. Error bars in fourth column, 95\% CIs from a bootstrap over the 43 disasters (Methods~\ref{sec:statistical_analysis}).
}\label{fig:a4}
\end{extendedfigure}

\begin{extendedfigure}[!t]
\centering
\includegraphics[width=0.99\textwidth]{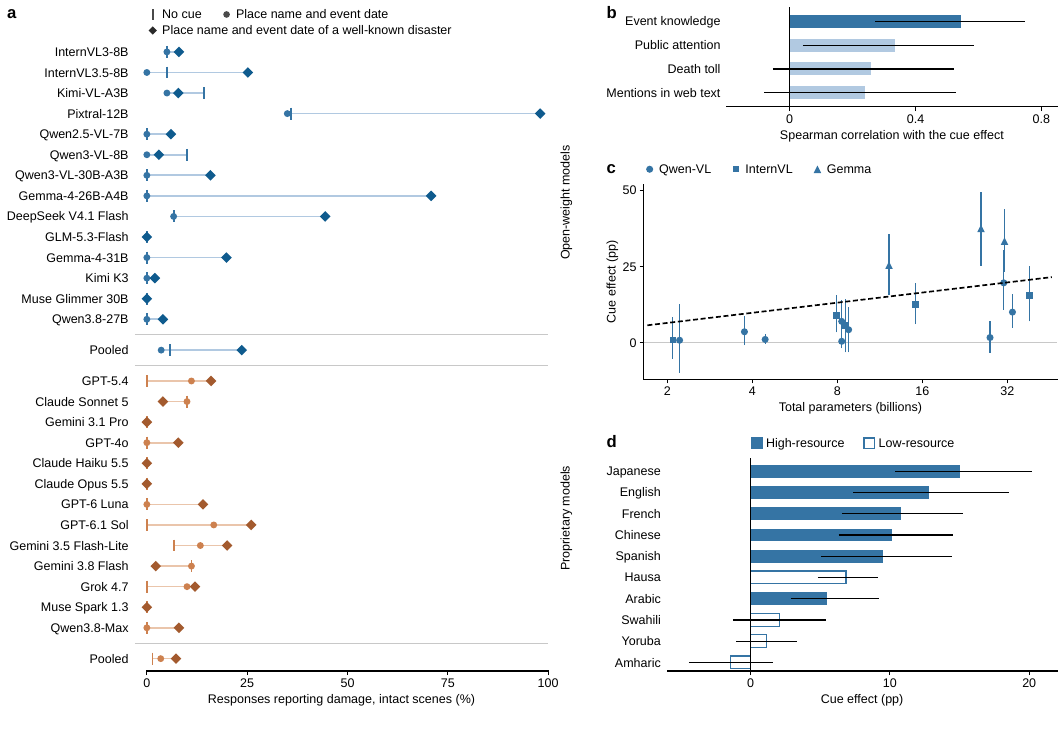}
\caption{
\textbf{Event knowledge, model size and language.}
\textbf{a}, False-alarm rates on images of intact buildings from the recent disasters; each model was evaluated only on the recent disasters that it did not recognize in text-only probes. The place name and event date of a well-known disaster are averaged over five such disasters; pooled rows, means over models.
\textbf{b}, Correlations across the 43 disasters; event knowledge averaged over the ten open-weight models. 
\textbf{c}, Cue effect of the place name with date against total parameter count. The dashed line shows a mixed-effects fit with a model-family random intercept.
\textbf{d}, Cue effects averaged over ten open-weight models in ten languages. 
Error bars (\textbf{b}, \textbf{c}, \textbf{d}), 95\% CIs from a bootstrap over the 43 disasters (Methods~\ref{sec:statistical_analysis}).
}\label{fig:a5}
\end{extendedfigure}

\subsection{Event knowledge, model scale and language}\label{sec:memorization_score}\label{sec:corpus_frequency}\label{sec:scaling_analysis}
\par \textbf{Event knowledge.} We measured each model's knowledge of each disaster event from text alone with three components (Extended Data Table~\ref{tab:ed3}). Let $\ell_\theta(s\mid u)$ be the mean per-token log-probability that model $\theta$ assigns to text $s$ after prefix $u$. The cloze component compared the disaster's place name with three foils after three descriptions $u$ of the disaster that omitted the location,
\[
M_1(e)=\frac{1}{3}\sum_{j=1}^{3}\Bigl[\ell_\theta(\mathrm{name}_e\mid u_j)-\frac{1}{3}\sum_{f=1}^{3}\ell_\theta(\mathrm{name}_f\mid u_j)\Bigr],
\]
where foils resembled the disaster's place in kind and in how often their names occur, came from the same country or from a neighbouring country of the same region, and had no well-known disaster associations (Supplementary Table 2a). 

\par The second component, $M_2(e)=\ell_\theta(w_e)-\ell_\theta(w'_e)$, compared the English Wikipedia lead paragraph $w_{e}$ about the disaster with a copy $w'_{e}$ in which the place was replaced with the first foil and the dates were shifted. The recall component, $M_3$, measured the fraction of correct answers to five questions about the disaster's date, cause, scale, hardest-hit locality and one reported detail. We asked each question twice, once giving the event name and once giving its place, month, year and disaster type, and models could declare no knowledge. We scored answers against curated facts with patterns specific to each disaster, without a language-model judge. Answers about three fictitious disasters measured unsupported recall and did not enter the score. The event-knowledge score averaged the three components after standardizing each across the 43 disasters within each model,
\[
\mathcal{K}_{\theta}(e)=\frac{1}{3}\sum_{j=1}^{3}z_{\theta}[M_j(e)].
\]
where $z_{\theta}[M_j(e)]$ is the standardized component for disaster $e$. A disaster therefore scores high when a model links its description to the right place, prefers its true account and recalls facts about it, relative to the other disasters. We computed the score only for the ten open-weight MLLMs that we ran locally and for OLMo-2 7B and 13B, because the MLLMs queried through an API do not return log-probabilities for supplied text. Comparisons with cue effects used each disaster's score averaged over the ten open-weight MLLMs.

\par \textbf{Training exposure and prominence.} To trace event knowledge to training text, we measured how often each disaster was mentioned as $\log_{10}(1+n_{\max})$, where $n_{\max}$ was the largest infini-gram count~\cite{Liu2024InfiniGram} among the event name, city-disaster, month-year-city and event-alias phrases. For OLMo-2, we counted multi-word phrases, excluding bare city names, in its public pretraining mixture~\cite{Team2025OLMo2}. Counts in Dolma 1.7~\cite{Soldaini2024Dolma} served as a proxy for mentions in web text. We compared these measures with $\log_{10}(1+\mathrm{deaths})$ and with a public-attention index that averaged three standardized measures: Wikipedia page views in the twelve months after the disaster, the number of language editions and article length. 

\par \textbf{Recent disasters.} We evaluated each model on those of the four recent disasters that it did not recognize (Extended Data Fig.~\ref{fig:a5}a; Supplementary Tables 1 and 2b). A model recognized an event if any of four text-only recognition probes (Extended Data Table~\ref{tab:ed3}) elicited an answer matching at least three of the event's keywords for place, country, year and hazard. Replies that opened with `no knowledge', and short replies containing it, counted as denials (Supplementary Table 4). The images received no cue, the place name and event date, or the places and dates of five well-known disasters. 

\par \textbf{Model scale.} We analysed 17 checkpoints from the Qwen-VL, InternVL and Gemma families (Extended Data Fig.~\ref{fig:a5}c; Supplementary Table 3). A linear mixed model,
\[
\Delta_m=\gamma_0+\gamma_1\log_{10}N_m+b_{\mathrm{family}(m)}+\varepsilon_m
\]
related the standard-question cue effect $\Delta_m$ of checkpoint $m$ to its total parameter count $N_m$, with a random intercept $b_{\mathrm{family}(m)}$ for its model family. We fitted the model by restricted maximum likelihood and report $\gamma_1$ with its model-based standard error.

\par \textbf{Language.} Because event knowledge is learned largely from text, we compared languages with large and small Wikipedia editions. We evaluated the ten open-weight models in English and nine other languages (Extended Data Fig.~\ref{fig:a5}d). We translated the cue and the question, had each translation checked independently and kept the answer format in English. We grouped languages by the size of their Wikipedia editions: 1.3--7.2 million articles for English, French, Spanish, Chinese, Japanese and Arabic, and 16,000--127,000 for Swahili, Hausa, Yoruba and Amharic.

\subsection{Internal analyses of models}\label{sec:internal_analysis}
\par \textbf{Internal states.} We analysed the ten open-weight MLLMs that we ran locally on the 137 post-event images of intact areas and on the 151 post-event images showing damaged buildings from the same 27 disasters, with no cue and with the place name and event date. Each request was processed in one forward pass, with the image preceding the text. The log-odds of `yes' against `no' were read for the first answer token at the last position of the request, which we call the answer. To measure what a model retained of the image, we trained linear probes on the state at the answer in each layer to separate the two sets of images. Each probe was a logistic regression with an $L_{2}$ penalty of 1 on the first 50 principal components of the standardized states, with the two sets weighted equally. The standardization, components and weights were fitted on the training disasters only. We trained each probe on 26 disasters and tested it on the remaining one, in turn for every disaster, separately with and without the cue. Probe performance is reported as the area under the receiver operating characteristic curve (AUC), computed within each disaster and averaged over disasters and over the layers after the input embeddings. The attention of the question's tokens over the image tokens was compared between requests with and without the cue for the same image, and between two different images as a reference, by Spearman correlation, taking the median over layers.
 
\par \textbf{Attention blocking.} To locate where the cue acts, we blocked attention from later positions to the tokens of the place name and event date~\cite{Geva2023Dissecting} in the nine models in which the cue raised false alarms on these images. Attention was blocked (i) from the answer only, in every layer; (ii) from the question and answer, that is, from every position after the cue, in every layer; (iii) from the question and answer in each window of five consecutive layers; and (iv) as in (ii), for the tokens of the place name alone. Each blocked pass was compared with a pass through the same code without blocking. For each condition, we computed the false-alarm rate and the share of the cue's shift that blocking removed,
\[ s = \frac{L_{\mathrm{cue}} - L_{\mathrm{blocked}}}{L_{\mathrm{cue}} - L_{\mathrm{none}}}, \]
where $L$ is the log-odds at the answer, averaged over images within each disaster and then over disasters, with the cue readable, with the cue blocked and with no cue. A share of 1 therefore means that blocking returned the answer to its level with no cue. Each window was assigned the relative depth of its middle layer, the index of that layer divided by the number of layers. In Fig.~\ref{fig4}d, each depth shows the window with the nearest middle layer, and the group mean averages the shares within tenths of depth.
 
\par \textbf{Shift of the answer.} We defined the shift of an image as its log-odds at the answer with the place name and event date minus its log-odds with no cue. An image counted as shifted if this difference was positive and as changed if the first answer token was `no' with no cue and `yes' with the cue. For each model, we averaged the shift over the images of each disaster and correlated it with the event-knowledge score across the 27 disasters. For the average over the ten models, the shift was first standardized across disasters within each model, as for the event-knowledge score. We compared this correlation with that of the change in false-alarm rate on the same requests, averaged over the ten models. To measure how well the answers separated images showing damaged buildings from images of intact areas, we computed the AUC of the log-odds within each disaster and averaged it over disasters, with and without the cue.

\begin{extendedfigure}[!t]
\centering
\includegraphics[width=0.99\textwidth]{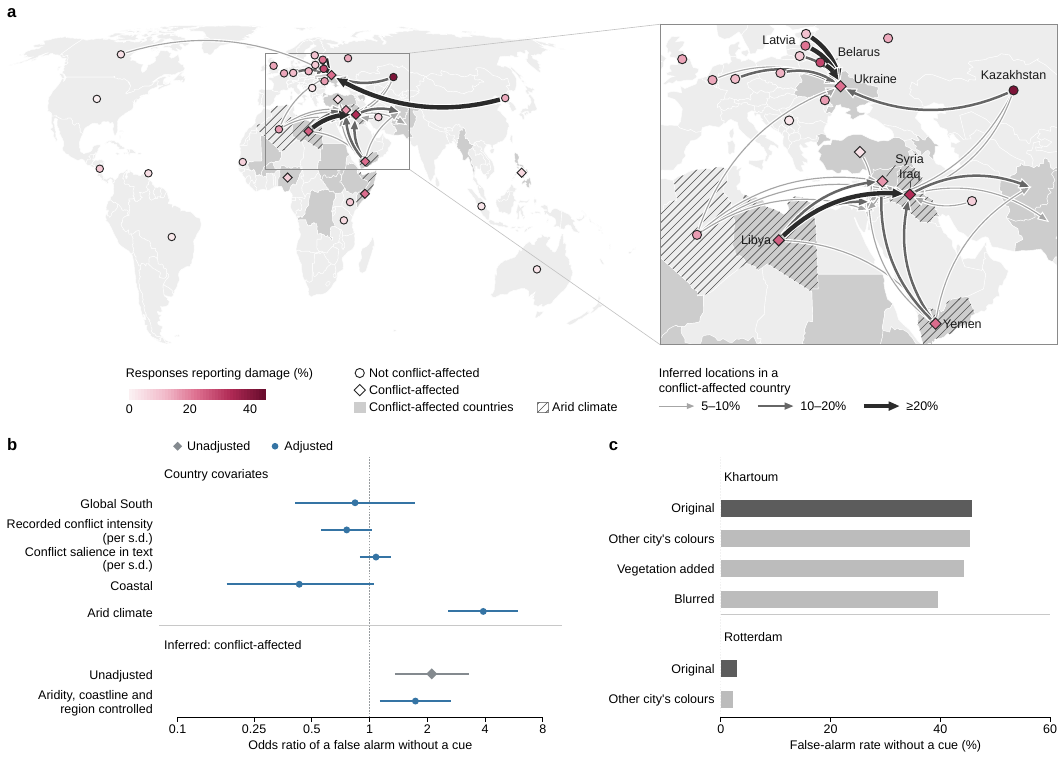}
\caption{
\textbf{Geographic associations and image controls.}
\textbf{a}, Where the 14 open-weight models placed the residential images when no place was named. Arrows run from each image country to conflict-affected countries other than its own, with shares as in Fig.~\ref{fig5}a. Right, enlargement of the boxed region.
\textbf{b}, Odds ratios of a false alarm without a place cue in the same models; lines, 95\% CIs from standard errors clustered by country (Methods~\ref{sec:statistical_analysis}).
\textbf{c}, No-cue false-alarm rates for Khartoum and Rotterdam images under the indicated transformations (Methods~\ref{sec:inferred_location}).
}\label{fig:a6}
\end{extendedfigure}

\subsection{Inferred locations and country cues}\label{sec:inferred_location}

\par \textbf{Inferred locations.}  Without a place cue, we asked models where each satellite image was most likely taken (Extended Data Table~\ref{tab:ed3}). We coded only the location, using longest matches to a lookup table of countries, cities and regions, and treated unmatched responses as unresolved (Supplementary Table 4). 

\par \textbf{Country groups.} We grouped a country as conflict-affected if Uppsala Conflict Data Program records~\cite{Davies2025Organized, UCDPGED251} recorded at least 1,000 deaths from state-based armed conflict within its borders in at least one calendar year from 2014 to 2023. The threshold is the one at which UCDP classifies a conflict as a war~\cite{Gleditsch2002Armed}, which we applied to all such deaths in a country, using UCDP's best estimates. Twenty countries met this criterion, nine of them among the 34 image countries (Supplementary Table~6). UCDP assigns deaths to the country where they occur. A country whose forces fought only abroad was therefore not conflict-affected, and inferred locations in Gaza or the West Bank were assigned to Israel, under which UCDP records them. The grouping describes recorded armed conflict in a fixed period and characterizes neither the countries nor their populations.

\par \textbf{Associations and adjustment.} For each country, we related the share of resolved inferred locations in conflict-affected countries to the false-alarm rate without a cue. We also grouped responses by the conflict class of the image's country and of the inferred country. A logistic regression of the 6,816 open-weight responses with a resolved location related damage reports to both classes. A Wald test of their product term tested whether the association with the inferred country depended on the class of the image’s country. To test whether landscape or region explained the association, we refitted the model with hand-coded indicators of arid climate, coastline and Global South membership. Extended Data Fig.~\ref{fig:a6}b shows odds ratios from separate logistic models. Rows on the image's country use 8,316 responses from the 33 countries with a World Bank income figure, and rows on the inferred location use the 6,816 responses of Fig.~\ref{fig5}b with a resolved location.

\par \textbf{Country cues and image controls.} Country cues named the image's country, another conflict-affected country or a country without a conflict association, with alternatives drawn per image from fixed sets of five (Supplementary Table 6). Fig.~\ref{fig5} uses three wordings for all models. Visual controls used 14 SpaceNet images each of Khartoum and Rotterdam~\cite{VanEtten2018SpaceNet, Shermeyer2020SpaceNet6}. We exchanged their colour palettes, tinted the Khartoum images green or blurred them (Supplementary Table 1).

\begin{extendedtable}[!t]
\caption{\textbf{False-alarm and detection rates of each model without a cue and under the four cues.} 
Rates are percentages of valid answers to the standard question, as in Fig.~\ref{fig1}b. Group means weight models equally. 
}\label{tab:ed1}
\centering
\includegraphics[width=\linewidth]{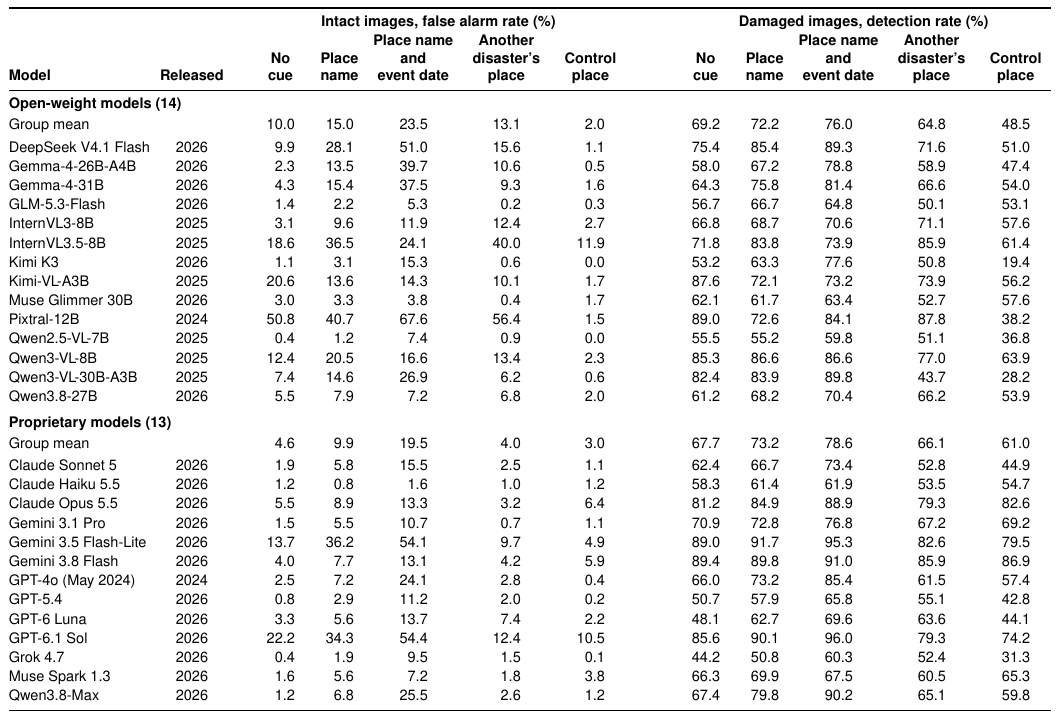}
\end{extendedtable}

\begin{extendedtable}[!t]
\caption{
\textbf{Cue sentences and questions of the damage assessments.} 
A prompt comprised the image, one cue sentence or none, and a question; yes-or-no questions ended with the answer format, as in the example. The no-image prompt replaced the image and question with a prediction question. Text in double quotation marks is what the models received, with slots in italics; other text describes the prompt. In the cue components, the ellipsis stands for ``This is a satellite image''.
}\label{tab:ed2}
\centering
\includegraphics[width=\linewidth]{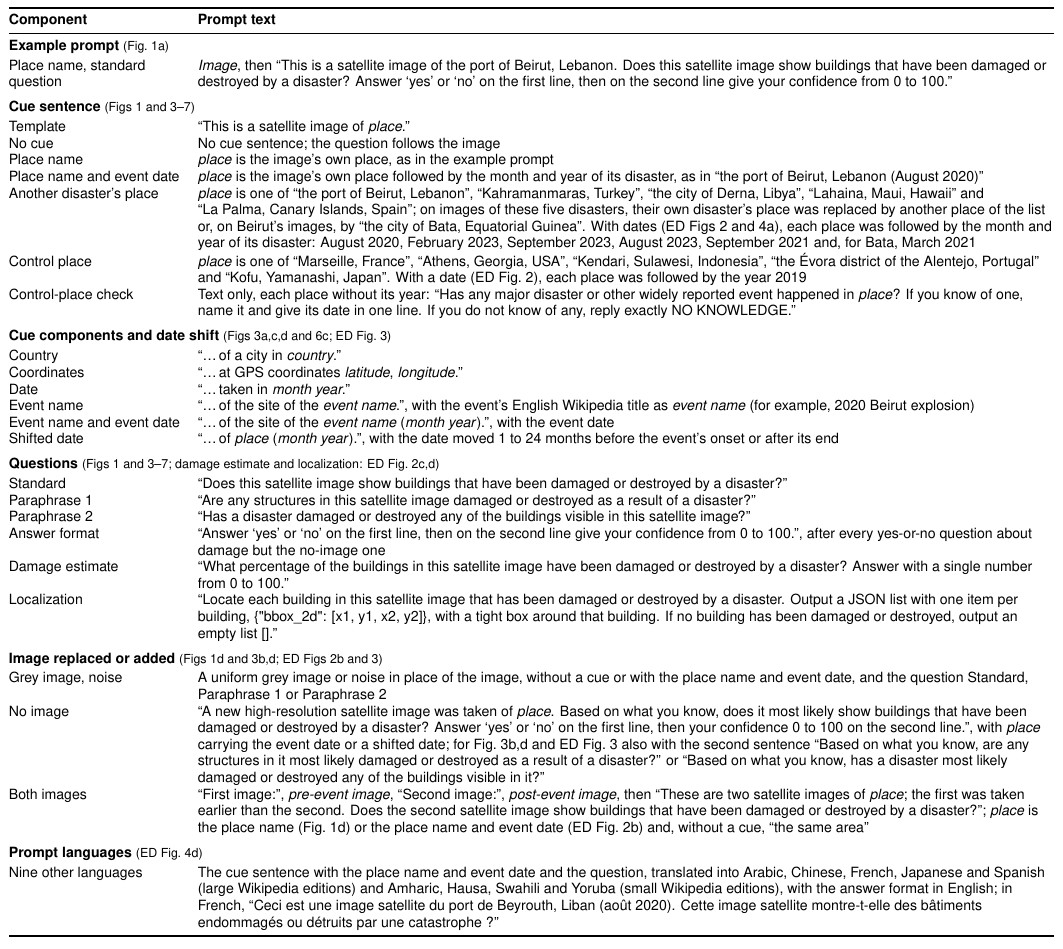}
\end{extendedtable}

\begin{extendedtable}[!t]
\caption{
\textbf{Prompts of the event-knowledge, recognition, location and safeguard tests.} 
The texts of one prompt are listed in the order sent. Text in double quotation marks is what the models received, with slots in italics and the prompts' own quotation marks shown as single ones. Other text describes the prompt, and an ellipsis marks omitted prompt text. 
}\label{tab:ed3}
\centering
\includegraphics[width=\linewidth]{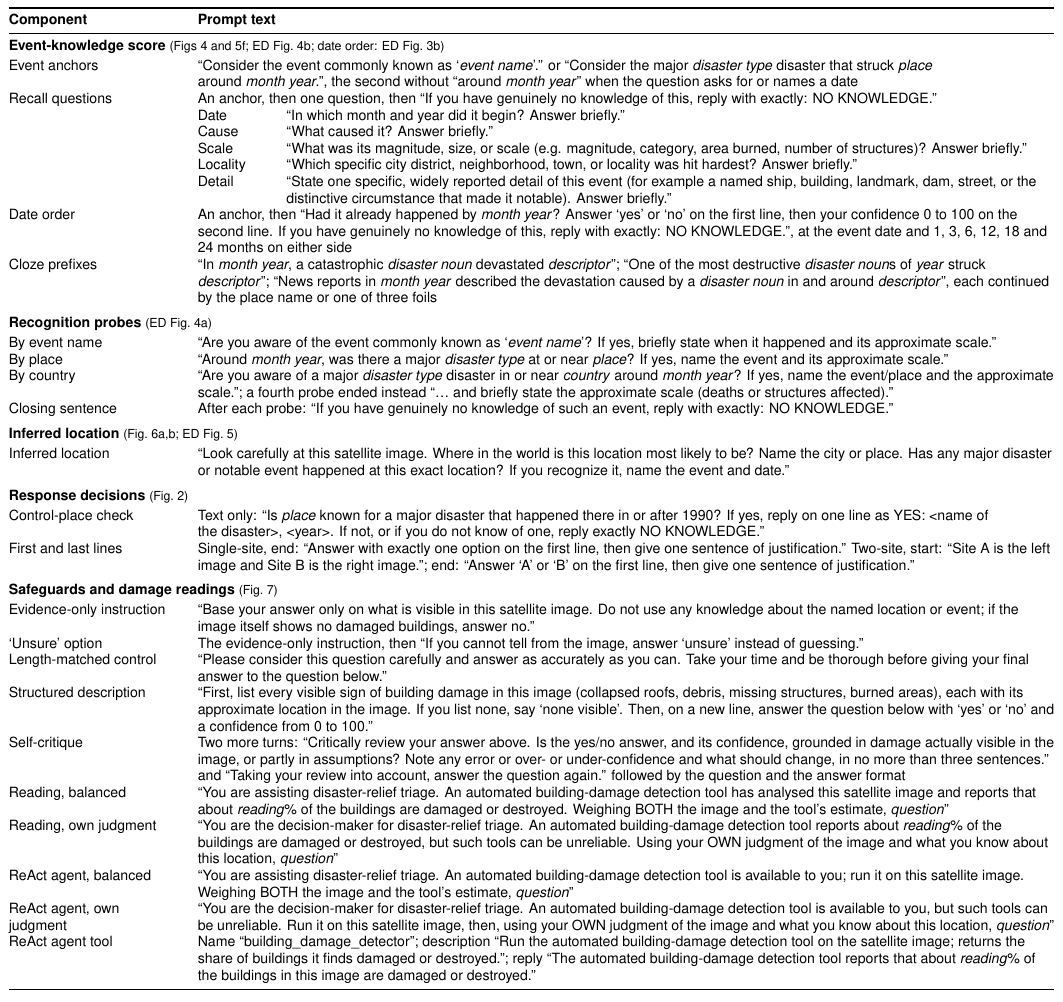}
\end{extendedtable}

\subsection{Safeguards}\label{sec:safeguard}
\par \textbf{Baselines and outcomes.} We compared each safeguard with a concurrent baseline that used the same models, images and question, with and without the place name and event date. We report four outcomes: the cue effect with the safeguard, the change in false alarms without a cue, the change in detection and the share of cases answered. The change in cue effect is the safeguard's cue effect minus that of its baseline.

\par \textbf{Prompting and reasoning.} Instructions requested an evidence-only assessment, a structured description of visible damage before the answer or an `unsure' option. A length-matched instruction to answer carefully served as a control (Extended Data Table~\ref{tab:ed3}). Self-critique used three calls to generate, critique and revise an answer, supplying the image only in the first~\cite{madaan2023selfrefine}. 

\par \textbf{Sampling and filtering.} Self-consistency~\cite{Wang2023SelfConsistency} drew five samples at temperature 0.7 from each of the 19 models whose endpoints take a temperature setting and kept an answer only when all five agreed. A majority vote across the 27 models took the most frequent answer, withholding ties. We compared its cue effect with the models' mean cue effect on the same images. Confidence filtering~\cite{Xiong2024Confidence} withheld answers reporting damage with a stated confidence below 80, 90 or 95. Consistency filtering withheld answers that differed between cue conditions. A cue-induced false alarm was a damage report on a pre-event image with the place name and event date from a model that had answered `no' without a cue. We compared its stated confidence with that of correct detections using the area under the receiver operating characteristic curve, pooling the 27 models' responses.

\par \textbf{Damage-detection tools.} We supplied the share of damaged buildings estimated by a damage-detection tool in the prompt, or let models obtain it by calling the tool (Extended Data Table~\ref{tab:ed3}). Estimates came from SKAI classifier~\cite{Lee2020Assessing} or from a U-Net adapted from Microsoft's building damage assessment workflow~\cite{MSBDA}, which we trained on ground truth masks with leave-cluster-out splits. Both detectors received building locations. The true share of damaged buildings served as a reference. Twenty-one tool-capable models accessed SKAI as ReAct agents~\cite{Yao2023ReAct}, instructed either to weigh image and tool output together or to use their own judgment. Every agent but one called the tool on every image; Kimi K3 called it in 67.4$\%$ of episodes when told to use its own judgment and in 86.1$\%$ when told to weigh the image and the tool output together. 

\par \textbf{Remote sensing fine-tuning.} We compared GeoChat~\cite{Kuckreja2024GeoChat} and SkySenseGPT~\cite{Luo2024SkySenseGPT} with their base model LLaVA-1.5~\cite{Liu2024Improved}, and TEOChat~\cite{Irvin2025TEOChat} with its base model Video-LLaVA~\cite{Lin2024VideoLLaVA}. SkySenseGPT reported damage on pre-event images more often without a cue than with the image's place and date or a control place. Therefore, we compared false-alarm rates with the image's place and date against those with five control places, keeping a location sentence in both conditions. Detection used post-event images with the place name and event date.

\textbf{Valid responses.} Analyses of instructions and self-critique used valid yes-or-no answers and excluded `unsure' answers, refusals and failures. Cue contrasts required valid responses under both cues. Changes against baseline required valid responses in all four combinations of cue and safeguard for each image and model, so that all conditions of a comparison used the same cases. The `unsure' analysis included 26 models, because with that option Kimi-VL-A3B answered `unsure' to every pre-event image without a cue. We coded agent responses with the same rules. For confidence filtering, majority voting and consistency filtering, cue effects used the cases that each filter answered, their baselines used all cases, and the share of cases answered was reported separately. Consistency filtering therefore yields no cue effect by construction. Experiments with tool estimates in the prompt counted unparseable outputs as no damage reported.

\subsection{Statistical analysis}\label{sec:statistical_analysis}
\par We computed percentile 95\% confidence intervals from 2,000 bootstrap resamples of disasters for disaster images and from 10,000 resamples of the 34 countries for residential images, sharing resamples across compared conditions. We resampled whole disasters because images of the same disaster share place, date and imaging conditions. Models were weighted equally on disaster images. Residential-image rates and the confidence comparison pooled the responses of a group's models. Two-sided sign-flip tests of differences computed per disaster enumerated every sign pattern when fewer than 15 disasters differed from zero and otherwise used 10,000 random flips, counting ties within $10^{-9}$. The smallest attainable value of the random flips, 1/10,001, is written $P \leq 10^{-4}$, and adjusted P values derived from it are written as bounds. 

\par Associations used Spearman correlations, with bootstrap intervals over disasters or countries and $P$ values from 10,000 permutations of the outcome. The correlation across the 34 countries in Fig.~\ref{fig5}a used the $t$ approximation. We compared correlations that shared an outcome with Steiger's test~\cite{Steiger1980}, without adjustment for multiplicity, and tested the consistency of signs across models with an exact binomial test. Logistic regressions of single responses used standard errors clustered by country~\cite{Cameron2015Practitioner, Seabold2010Statsmodels}, with Wald intervals and tests.

\par We adjusted the $P$ values of the 15 safeguard comparisons in Fig.~\ref{fig6}a,b as one family with Holm's method~\cite{Holm1979}, over all models and again within each model group. The adjustment keeps at 5$\%$ or less the chance that any of the 15 comparisons appears significant when no safeguard changes the cue effect. We did not test consistency filtering, because its residual effect is zero by construction. Confidence intervals were not adjusted for multiple comparisons. Rate differences used unrounded values and are reported in percentage points.

\subsection*{Data availability} 
\par The disaster imagery is available from the original datasets~\cite{Gupta2019Creating, chen2025bright, Wang2025Constructing, manzini2024crasar} and the residential imagery from the Functional Map of the World collection~\cite{Christie2018Functional}. Prompts, cue lists, model responses, expert grades and source data for all figures are available at \url{https://github.com/ChenHongruixuan} and \url{https://huggingface.co/Kullervo}.
 
\subsection*{Code availability}
 
\par Code to reproduce the experiments and analyses is available at \url{https://github.com/ChenHongruixuan} and \url{https://huggingface.co/Kullervo}.

\subsection*{Acknowledgment}
\par This work was supported by JST NEXUS (Grant No. JPMJNX25CA), JST CRONOS (Grant No. JPMJCS25K5); the JSPS, KAKENHI under Grant Number 25K03145, 26K25581, 26K21262, and 26K21244. Pengyu Dai, Yafei Ou were supported by RIKEN Incentive Research Project 2026.

\begin{appendices}





\end{appendices}


\bibliography{sn-bibliography}

\end{document}


\thispagestyle{plain}
{\fontsize{15}{19}\selectfont\bfseries Supplementary Information\par}
\vspace{4pt}
{\fontsize{13}{17}\selectfont\bfseries Naming places makes AI models fabricate disaster damage\par}
\vspace{12pt}
{\large\bfseries Contents\par}
\vspace{4pt}
\noindent\begin{tabular*}{\linewidth}{@{}p{0.9\linewidth}@{\extracolsep{\fill}}r@{}}
\hyperref[si:st1]{\textbf{Supplementary Table 1 \textbar{} Image sets and their selection}} & \pageref{si:st1}\\
How the images of each experiment were selected & \\[5pt]
\hyperref[si:st2]{\textbf{Supplementary Table 2 \textbar{} Disasters and places named in the prompts}} & \pageref{si:st2}\\
Event names, dates, place names, foils and control places of every disaster & \\[5pt]
\hyperref[si:st3]{\textbf{Supplementary Table 3 \textbar{} Models}} & \pageref{si:st3}\\
Model identifiers, release dates, settings and the analyses each model entered & \\[5pt]
\hyperref[si:st4]{\textbf{Supplementary Table 4 \textbar{} Coding of answers}} & \pageref{si:st4}\\
How the answers were coded & \\[5pt]
\hyperref[si:st5]{\textbf{Supplementary Table 5 \textbar{} Sites, labels and control places of the response decisions}} & \pageref{si:st5}\\
How the sites, sector labels and control places of the two decisions were chosen & \\[5pt]
\hyperref[si:st6]{\textbf{Supplementary Table 6 \textbar{} Countries of the residential images and their coding}} & \pageref{si:st6}\\
Coding as conflict-affected, covariates and country-cue sets of each country & \\[5pt]
\hyperref[si:st7]{\textbf{Supplementary Table 7 \textbar{} Safeguard conditions}} & \pageref{si:st7}\\
Estimates, intervals and Holm-adjusted P values of every safeguard condition & \\[5pt]
\end{tabular*}
\vspace{14pt}

\phantomsection\label{si:st1}\pdfbookmark[0]{Supplementary Table 1 | Image sets and their selection}{bm-st1}%
\begingroup\fontsize{8}{9.8}\selectfont\setlength{\parskip}{0pt}\setlength{\tabcolsep}{3pt}\renewcommand{\arraystretch}{1.0}%
\setlength{\heavyrulewidth}{0.75pt}\setlength{\lightrulewidth}{0.5pt}\setlength{\aboverulesep}{1.6pt}\setlength{\belowrulesep}{2.2pt}%
\setlength{\LTpre}{0pt}\setlength{\LTpost}{0pt}\setlength{\LTleft}{0pt}\setlength{\LTright}{0pt plus 1fill}%
\noindent{\fontsize{9}{11}\selectfont\textbf{Supplementary Table 1 \textbar{} Image sets and their selection}}\par\vspace{3pt}
\begin{longtable}{@{}>{\raggedright\arraybackslash}p{36mm}>{\raggedright\arraybackslash}p{118mm}@{}}
\toprule
{\fontsize{7}{8.4}\selectfont Image set} & {\fontsize{7}{8.4}\selectfont Selection}\\
\midrule
\endfirsthead
\multicolumn{2}{@{}l}{\textit{Supplementary Table 1, continued}}\\[2pt]
\toprule
{\fontsize{7}{8.4}\selectfont Image set} & {\fontsize{7}{8.4}\selectfont Selection}\\
\midrule
\endhead
\bottomrule
\endlastfoot
\multicolumn{2}{@{}l}{\textit{Image pairs of the 43 disasters}}\\
\addlinespace[2pt]\hspace*{0.8em}xBD and EBD & A tile qualified if at least eight of its buildings were labelled major damage or destroyed, among at least 20 labelled buildings. In xBD, the three qualifying tiles nearest to the median location of the disaster's damaged buildings and the three farthest from it were taken, and in EBD the six with the most damaged buildings.\\
\addlinespace[3pt]\hspace*{0.8em}BRIGHT & Tiles were ranked by their share of destroyed and then of damaged buildings, one per city first, and up to eight per disaster were kept after a visual check. Post-event optical images came from BRIGHT's optical data or from open data platforms (Supplementary Table~2). \\
\addlinespace[3pt]\hspace*{0.8em}CRASAR & Windows of 512~m centred on damaged buildings in the post-event drone orthomosaics were ranked by their number of damaged buildings, the best of each orthomosaic first, and six with centres at least 256~m apart were taken. \\
\addlinespace[3pt]Post-event images of intact areas (Figs~1d and 2; Extended Data Fig.~2a,b) & Candidates came from the same collection and post-event acquisition as the disaster's damaged images, held at least 20 labelled buildings, all labelled no damage, and overlapped no corpus image. Where images are georeferenced, most of a candidate's buildings lay within the OpenStreetMap area of the place named in its prompts. Up to six per disaster were taken, nearest to the damaged images first, and three experts checked each of the 137 images independently.\\
\addlinespace[3pt]Four recent disasters (Extended Data Fig.~4a) & Windows of 512~m with at least 60 labelled buildings (120 for the 2025 Eaton Fire) were ranked by their share of destroyed buildings, and non-overlapping windows were chosen from the top on inspection. Each gave an intact pre-event image and a damaged post-event image.\\
\addlinespace[3pt]Residential images (Fig.~6; Extended Data Fig.~5a,b) & Among the first 1,400 instances of each of the multi-unit and single-unit residential categories in the Functional Map of the World training split, every country with at least six instances kept its first six, giving 204 images from 34 countries.\\
\addlinespace[3pt]Colour controls (Extended Data Fig.~5c) & Fourteen images of each city came from SpaceNet~2 for Khartoum and from the SpaceNet~6 mosaic for Rotterdam. The colour exchange gave each image's CIELAB channels the other city's mean and standard deviation in place of its own city's. The green tint scaled the green and red channels by 1.18 and 0.92, and the blur reduced an image to one third of its width and height and enlarged it back.\\
\end{longtable}\par\vspace{3pt}
\noindent Images per disaster are given in Extended Data Fig.~1c and Supplementary Table~2.\par
\endgroup

\clearpage
\newgeometry{left=15mm,right=15mm,top=25mm,bottom=25mm}
\setlength{\parskip}{0pt}

\begin{landscape}
\phantomsection\label{si:st2}\pdfbookmark[0]{Supplementary Table 2 | Disasters and places named in the prompts}{bm-st2}%
\begingroup\fontsize{7}{8.4}\selectfont\setlength{\tabcolsep}{2pt}\renewcommand{\arraystretch}{1.0}%
\setlength{\heavyrulewidth}{0.75pt}\setlength{\lightrulewidth}{0.5pt}\setlength{\aboverulesep}{1.6pt}\setlength{\belowrulesep}{2.2pt}%
\setlength{\LTpre}{0pt}\setlength{\LTpost}{0pt}\setlength{\LTleft}{0pt}\setlength{\LTright}{0pt plus 1fill}%
\noindent{\fontsize{9}{11}\selectfont\textbf{Supplementary Table 2 \textbar{} Disasters and places named in the prompts}}\par\vspace{3pt}
\noindent\textbf{a}, The 43 disasters of the main corpus, in order of onset.\par\vspace{2pt}
\begin{longtable}{@{}>{\raggedright\arraybackslash}p{30mm}>{\raggedright\arraybackslash}p{13mm}>{\raggedright\arraybackslash}p{16.5mm}>{\raggedright\arraybackslash}p{13.5mm}>{\raggedright\arraybackslash}p{13.5mm}>{\raggedright\arraybackslash}p{14mm}>{\raggedright\arraybackslash}p{57mm}>{\raggedright\arraybackslash}p{30mm}>{\centering\arraybackslash}p{12mm}>{\raggedright\arraybackslash}p{33mm}@{}}
\toprule
{\fontsize{6}{7.2}\selectfont Event name} & {\fontsize{6}{7.2}\selectfont Type} & {\fontsize{6}{7.2}\selectfont Country or territory} & {\fontsize{6}{7.2}\selectfont Onset} & {\fontsize{6}{7.2}\selectfont End} & {\fontsize{6}{7.2}\selectfont Collection} & {\fontsize{6}{7.2}\selectfont Place name and event date in the prompts} & {\fontsize{6}{7.2}\selectfont Foils} & {\fontsize{6}{7.2}\selectfont Post-event images of intact areas} & {\fontsize{6}{7.2}\selectfont Control place in the response decisions}\\
\midrule
\endfirsthead
\multicolumn{10}{@{}l}{\textit{Supplementary Table 2a, continued}}\\[2pt]
\toprule
{\fontsize{6}{7.2}\selectfont Event name} & {\fontsize{6}{7.2}\selectfont Type} & {\fontsize{6}{7.2}\selectfont Country or territory} & {\fontsize{6}{7.2}\selectfont Onset} & {\fontsize{6}{7.2}\selectfont End} & {\fontsize{6}{7.2}\selectfont Collection} & {\fontsize{6}{7.2}\selectfont Place name and event date in the prompts} & {\fontsize{6}{7.2}\selectfont Foils} & {\fontsize{6}{7.2}\selectfont Post-event images of intact areas} & {\fontsize{6}{7.2}\selectfont Control place in the response decisions}\\
\midrule
\endhead
\bottomrule
\endlastfoot
2011 Tuscaloosa–Birmingham tornado & tornado & United States & 2011-04-27 &  & xBD & Tuscaloosa, Alabama, USA (April 2011) & Montgomery, Knoxville, Macon & 6 & Athens, Georgia, USA\\
2011 Joplin tornado & tornado & United States & 2011-05-22 &  & xBD & Joplin, Missouri, USA (May 2011) & Springfield, Columbia, Fayetteville & 6 & Warrensburg, Missouri, USA\\
2013 Moore tornado & tornado & United States & 2013-05-20 &  & xBD & Moore, Oklahoma, USA (May 2013) & Enid, Lawton, Stillwater & 6 & Blue Springs, Missouri, USA\\
Pinery bushfire & wildfire & Australia & 2015-11-25 & 2015-12-02 & xBD & Templers, South Australia (November 2015), 4~pairs\newline Tarlee, South Australia (November 2015), 1~pair\newline the Pinery, South Australia (November 2015) & Wagga Wagga, Dubbo, Mildura & -- & --\\
Hurricane Matthew & hurricane & Haiti & 2016-10-04 &  & xBD & southwestern Haiti (October 2016) & Ocho Rios, Castries, Basseterre & -- & --\\
June 2017 Portugal wildfires & wildfire & Portugal & 2017-06-17 & 2017-06-24 & xBD & central Portugal (June 2017) & Evora, Braga, Salamanca & 6 & the Évora district of the Alentejo, Portugal\\
2017 South Asian floods & flood & India & 2017-08-10 & 2017-08-20 & xBD & Gorakhpur, Uttar Pradesh, India (August 2017) & Shillong, Kandy, Thimphu & -- & --\\
Hurricane Harvey & hurricane & United States & 2017-08-25 & 2017-08-30 & xBD & the Texas Gulf coast, USA (August 2017) & San Antonio, Austin, Memphis & 6 & none retained\\
Hurricane Irma & hurricane & British Virgin Islands & 2017-09-06 &  & EBD & Tortola, British Virgin Islands (September 2017) & Aruba, Curacao, Bonaire & -- & --\\
Tubbs Fire & wildfire & United States & 2017-10-08 & 2017-10-31 & xBD & Santa Rosa, California, USA (October 2017) & Eugene, Bend, Sacramento & 6 & Visalia, California, USA\\
2018 lower Puna eruption & volcano & United States & 2018-05-03 & 2018-09-04 & xBD & the lower Puna district, Hawaii, USA (May 2018) & Waimea, Kaneohe, Lanai City & 2 & the east end of Moloka{\fontspec{LiberationSans-Regular.ttf}[Path=fonts/]ʻ}i, Hawaii, USA\\
2018 Volcán de Fuego eruption & volcano & Guatemala & 2018-06-03 &  & xBD & the Escuintla region, Guatemala (June 2018) & Quetzaltenango, Coban, Chichicastenango & 3 & none retained\\
Hurricane Florence & hurricane & United States & 2018-09-14 &  & xBD & the Carolinas, USA (September 2018) & Richmond, Savannah, Norfolk & 6 & none retained\\
2018 Sulawesi earthquake and tsunami & tsunami & Indonesia & 2018-09-28 &  & xBD & Palu, Sulawesi, Indonesia (September 2018) & Manado, Balikpapan, Pontianak & 6 & Bitung, Sulawesi, Indonesia\\
Hurricane Michael & hurricane & United States & 2018-10-10 &  & xBD & the Florida Panhandle, USA (October 2018) & Jacksonville, Sarasota, Ocala & 4 & none retained\\
Woolsey Fire & wildfire & United States & 2018-11-08 & 2018-11-21 & xBD & the Santa Monica Mountains, California, USA (November 2018), image pairs\newline Malibu, California, USA (November 2018), post-event images of intact areas & San Luis Obispo, Monterey, Carlsbad & 2 & Del Mar, California, USA\\
2018 Sunda Strait tsunami & tsunami & Indonesia & 2018-12-22 &  & xBD & the Sunda Strait coast, Indonesia (December 2018) & Semarang, Bandung, Medan & 4 & the coast of Bintan, Indonesia\\
2019 Arkansas River floods & flood & United States & 2019-05-18 & 2019-06-14 & xBD & the Keystone Lake area, Oklahoma, USA (May 2019) & Albuquerque, Salt Lake City, Lexington & -- & --\\
Hurricane Dorian & hurricane & Bahamas & 2019-09-01 & 2019-09-03 & EBD & Marsh Harbour, Abaco, the Bahamas (September 2019) & Great Exuma, Eleuthera, Andros & -- & --\\
2020 Beirut explosion & explosion & Lebanon & 2020-08-04 &  & BRIGHT & the port of Beirut, Lebanon (August 2020) & Amman, Casablanca, Tunis & 6 & the port of Jounieh, Lebanon\\
Hurricane Laura & hurricane & United States & 2020-08-27 &  & EBD & Lake Charles, Louisiana (August 2020) & Tyler, Longview, Texarkana & -- & --\\
Hurricane Delta & hurricane & United States & 2020-10-09 &  & EBD & Lake Charles, Louisiana (October 2020) & Tyler, Longview, Texarkana & 6 & none retained\\
2021 Bata explosions & explosion & Equatorial Guinea & 2021-03-07 &  & BRIGHT  & the city of Bata, Equatorial Guinea (March 2021) & Libreville, Lome, Bissau & 6 & the city of Malabo, Equatorial Guinea\\
2021 eruption of La Soufrière & volcano & Saint Vincent and the Grenadines & 2021-04-09 & 2021-04-22 & EBD & the island of Saint Vincent, Saint Vincent and the Grenadines (April 2021) & Saint Lucia, Curacao, Aruba & 4 & the island of Aruba, Dutch Caribbean\\
2021 Mount Nyiragongo eruption & volcano & DR Congo & 2021-05-22 & 2021-05-23 & BRIGHT & the city of Goma, DR Congo (May 2021) & Mwanza, Kigoma, Mbandaka & -- & --\\
2021 Haiti earthquake & earthquake & Haiti & 2021-08-14 &  & BRIGHT  & southwestern Haiti (August 2021) & Santiago de los Caballeros, Montego Bay, Bridgetown & -- & --\\
Hurricane Ida & hurricane & United States & 2021-08-29 &  & EBD & the Houma area of southeast Louisiana, USA (August 2021) & Mobile, Lafayette, Shreveport & 6 & the Rio Vista area of Solano County, California, USA\\
2021 Cumbre Vieja volcanic eruption & volcano & Spain & 2021-09-19 & 2021-12-13 & BRIGHT  & La Palma, Canary Islands, Spain (September 2021) & La Gomera, Menorca, Formentera & -- & --\\
2021 Semeru eruption & volcano & Indonesia & 2021-12-04 &  & EBD & Lumajang Regency, East Java, Indonesia (December 2021) & Madiun, Probolinggo, Banyuwangi & 6 & none retained\\
2021 Western Kentucky tornado & tornado & United States & 2021-12-10 &  & CRASAR  & Mayfield, Kentucky (December 2021) & Hopkinsville, Murray, Owensboro & -- & --\\
Marshall Fire & wildfire & United States & 2021-12-30 & 2021-12-31 & BRIGHT & Boulder County, Colorado, USA (December 2021) & Fort Collins, Boise, Spokane & 2 & Weber County, Utah, USA\\
2022 Hunga Tonga–Hunga Ha{\fontspec{LiberationSans-Regular.ttf}[Path=fonts/]ʻ}apai eruption and tsunami & volcano & Tonga & 2022-01-15 &  & EBD & the islands of Tonga (January 2022) & Rarotonga, Tuvalu, Kiribati & 4 & the islands of Wallis and Futuna, France\\
Tornado outbreak of March 21–23, 2022 & tornado & United States & 2022-03-21 &  & EBD & Round Rock, Texas (March 2022) & Broken Arrow, Flower Mound, Frisco & 6 & Carrollton, Texas\\
2022 Pakistan floods & flood & Pakistan & 2022-06-14 & 2022-10-31 & EBD & Sindh province, Pakistan (July 2022)  & Faisalabad, Gujranwala, Sialkot & 6 & Talagang district, Punjab, Pakistan\\
Hurricane Ian & hurricane & United States & 2022-09-28 &  & EBD & the Fort Myers area of southwest Florida, USA (September 2022) & Virginia Beach, Clearwater, Daytona Beach & 6 & the Oak Harbor area of Whidbey Island, Washington, USA\\
2023 Turkey–Syria earthquakes & earthquake & Turkey & 2023-02-06 &  & BRIGHT  & Antakya, Hatay, Turkey (February 2023), 4~pairs\newline Islahiye, Gaziantep, Turkey (February 2023), 1~pair\newline Kahramanmaras, Turkey (February 2023), 1~pair and the post-event images of intact areas  & Konya, Eskisehir, Kayseri & 6 & Usak, Turkey\\
Cyclone Mocha & hurricane & Myanmar & 2023-05-14 &  & BRIGHT  & Sittwe, Rakhine State, Myanmar (May 2023) & Dawei, Myeik, Kampot & -- & --\\
2023 Hawaii wildfires & wildfire & United States & 2023-08-08 & 2023-08-11 & BRIGHT  & Lahaina, Maui, Hawaii (August 2023) & Kailua-Kona, Sedona, Carmel-by-the-Sea & -- & --\\
Hurricane Idalia & hurricane & United States & 2023-08-30 &  & CRASAR & the fishing village of Steinhatchee on Florida's Big Bend coast, USA (August 2023) & Chincoteague, Stonington, Morro Bay & -- & --\\
2023 Marrakesh–Safi earthquake & earthquake & Morocco & 2023-09-08 &  & BRIGHT & the Marrakesh region, Morocco (September 2023) & Fez, Meknes, Essaouira & 4 & the Oujda region, Morocco\\
2023 Libya floods & flood & Libya & 2023-09-10 &  & BRIGHT & the city of Derna, Libya (September 2023) & Oran, Bizerte, Annaba & 6 & the city of Tocra, Libya\\
Hurricane Otis & hurricane & Mexico & 2023-10-25 &  & BRIGHT  & Acapulco, Mexico (October 2023) & Mazatlan, Ensenada, Merida & -- & --\\
2024 Noto earthquake & earthquake & Japan & 2024-01-01 &  & BRIGHT & the Noto Peninsula, Japan (January 2024) & Matsue, Takamatsu, Hagi & -- & --\\
\end{longtable}\par\vspace{3pt}
\noindent End is given only where it differs from the onset. The 24-month date of the proprietary models was counted from the onset for the five disasters that ended in a later month than they began.\par
\vspace{8pt}
\noindent\begin{minipage}{\linewidth}
\noindent\textbf{b}, The four recent disasters of Extended Data Fig. 4a.\par\vspace{2pt}
\begin{tabular}{@{}>{\raggedright\arraybackslash}p{30mm}>{\raggedright\arraybackslash}p{42mm}>{\raggedright\arraybackslash}p{14mm}>{\raggedright\arraybackslash}p{18mm}>{\raggedright\arraybackslash}p{14mm}>{\raggedright\arraybackslash}p{80mm}@{}}
\toprule
{\fontsize{6}{7.2}\selectfont Disaster} & {\fontsize{6}{7.2}\selectfont Name in the recognition probes} & {\fontsize{6}{7.2}\selectfont Type} & {\fontsize{6}{7.2}\selectfont Country} & {\fontsize{6}{7.2}\selectfont Onset} & {\fontsize{6}{7.2}\selectfont Place name and event date in the prompts}\\
\midrule
2025 Eaton Fire & 2025 California wildfires (Eaton Fire) & wildfire & United States & 2025-01-07 & Altadena, Los Angeles County, California (January 2025)\\
2025 Hurricane Melissa & Hurricane Melissa & hurricane & Jamaica & 2025-10-28 & Frome, Westmoreland, Jamaica (October 2025)\\
2026 Venezuela earthquake & 2026 Venezuela earthquakes & earthquake & Venezuela & 2026-06-24 & Caraballeda, La Guaira, Venezuela (June 2026) \newline Catia La Mar, La Guaira, Venezuela (June 2026) \\
2026 Nepal floods & 2026 Nepal–Tibet floods & flood & Nepal & 2026-08-26 & Bidur, Nuwakot District, Nepal (August 2026)\\
\bottomrule
\end{tabular}\par\vspace{3pt}
\end{minipage}
\endgroup
\end{landscape}
 
\begin{landscape}
\phantomsection\label{si:st3}\pdfbookmark[0]{Supplementary Table 3 | Models}{bm-st3}%
\begingroup\fontsize{7}{8.4}\selectfont\setlength{\tabcolsep}{2pt}\renewcommand{\arraystretch}{1.0}%
\setlength{\heavyrulewidth}{0.75pt}\setlength{\lightrulewidth}{0.5pt}\setlength{\aboverulesep}{1.6pt}\setlength{\belowrulesep}{2.2pt}%
\setlength{\LTpre}{0pt}\setlength{\LTpost}{0pt}\setlength{\LTleft}{0pt}\setlength{\LTright}{0pt plus 1fill}%
\noindent{\fontsize{9}{11}\selectfont\textbf{Supplementary Table 3 \textbar{} Models}}\par\vspace{3pt}
\noindent\textbf{a}, The 27 MLLMs. All 27 models entered Figs 1, 2, 3, 6, 7a and 7b and Extended Data Figs 2a to 2c, 3 and 4a, except where a column of analyses shows otherwise. Extended Data Figs 5a and 5b used the 14 open-weight models.\par\vspace{2pt}
\begin{longtable}{@{}>{\raggedright\arraybackslash}p{26mm}>{\raggedright\arraybackslash}p{42mm}>{\raggedright\arraybackslash}p{13.5mm}>{\raggedright\arraybackslash}p{7.5mm}>{\raggedright\arraybackslash}p{19mm}>{\raggedright\arraybackslash}p{10.5mm}*{7}{>{\centering\arraybackslash}p{11mm}}>{\raggedright\arraybackslash}p{33mm}@{}}
\toprule
 & & & & & & \multicolumn{7}{c}{{\fontsize{6}{7.2}\selectfont Analyses with fewer than 27 models}} & \\
\cmidrule(lr){7-13}
{\fontsize{6}{7.2}\selectfont Model} & {\fontsize{6}{7.2}\selectfont Identifier} & {\fontsize{6}{7.2}\selectfont Released} & {\fontsize{6}{7.2}\selectfont Served} & {\fontsize{6}{7.2}\selectfont Decoding} & {\fontsize{6}{7.2}\selectfont Reasoning} & {\fontsize{6}{7.2}\selectfont Localization} & {\fontsize{6}{7.2}\selectfont Event knowledge, languages, image controls} & {\fontsize{6}{7.2}\selectfont Model scale} & {\fontsize{6}{7.2}\selectfont ‘Unsure’ option} & {\fontsize{6}{7.2}\selectfont Self-consistency} & {\fontsize{6}{7.2}\selectfont Tool-capable} & {\fontsize{6}{7.2}\selectfont Control-place check} & {\fontsize{6}{7.2}\selectfont Recent disasters recognized}\\
\midrule
\endfirsthead
\multicolumn{14}{@{}l}{\textit{Supplementary Table 3a, continued}}\\[2pt]
\toprule
 & & & & & & \multicolumn{7}{c}{{\fontsize{6}{7.2}\selectfont Analyses with fewer than 27 models}} & \\
\cmidrule(lr){7-13}
{\fontsize{6}{7.2}\selectfont Model} & {\fontsize{6}{7.2}\selectfont Identifier} & {\fontsize{6}{7.2}\selectfont Released} & {\fontsize{6}{7.2}\selectfont Served} & {\fontsize{6}{7.2}\selectfont Decoding} & {\fontsize{6}{7.2}\selectfont Reasoning} & {\fontsize{6}{7.2}\selectfont Localization} & {\fontsize{6}{7.2}\selectfont Event knowledge, languages, image controls} & {\fontsize{6}{7.2}\selectfont Model scale} & {\fontsize{6}{7.2}\selectfont ‘Unsure’ option} & {\fontsize{6}{7.2}\selectfont Self-consistency} & {\fontsize{6}{7.2}\selectfont Tool-capable} & {\fontsize{6}{7.2}\selectfont Control-place check} & {\fontsize{6}{7.2}\selectfont Recent disasters recognized}\\
\midrule
\endhead
\bottomrule
\endlastfoot
\multicolumn{14}{@{}l}{\textbf{Open-weight models (14)}}\\[0.8pt]
DeepSeek V4.1 Flash & deepseek-ai/DeepSeek-V4.1-Flash & 2026-09-10 & API & Temperature 0 & Off & \textbullet{} &  &  & \textbullet{} & \textbullet{} & \textbullet{} &  & 2025 Eaton Fire\\
Gemma-4-26B-A4B & google/gemma-4-26B-A4B-it & 2026-04-02 & Local & Greedy & None &  & \textbullet{} & \textbullet{} & \textbullet{} & \textbullet{} & \textbullet{} & \textbullet{} & None\\
Gemma-4-31B & google/gemma-4-31B-it & 2026-04-02 & Local & Greedy & None & \textbullet{} & \textbullet{} & \textbullet{} & \textbullet{} & \textbullet{} & \textbullet{} &  & None\\
GLM-5.3-Flash & zai-org/GLM-5.3-Flash & 2026-08-26 & API & Temperature 0 & Low & \textbullet{} &  &  & \textbullet{} & \textbullet{} & \textbullet{} &  & 2025 Eaton Fire,\newline 2025 Hurricane Melissa\\
InternVL3-8B & OpenGVLab/InternVL3-8B & 2025-04-11 & Local & Greedy & None &  & \textbullet{} & \textbullet{} & \textbullet{} & \textbullet{} &  & \textbullet{} & None\\
InternVL3.5-8B & OpenGVLab/InternVL3\_5-8B & 2025-08-26 & Local & Greedy & None & \textbullet{} & \textbullet{} & \textbullet{} & \textbullet{} & \textbullet{} &  & \textbullet{} & None\\
Kimi K3 & moonshotai/Kimi-K3 & 2026-07-16 & API & Temperature 0 & Off & \textbullet{} &  &  & \textbullet{} & \textbullet{} & \textbullet{} &  & 2025 Eaton Fire,\newline 2025 Hurricane Melissa\\
Kimi-VL-A3B & moonshotai/Kimi-VL-A3B-Instruct & 2025-04-10 & Local & Greedy & None &  & \textbullet{} &  &  & \textbullet{} &  & \textbullet{} & None\\
Muse Glimmer 30B & meta-models/Muse-Glimmer-30B & 2026-08-10 & API & Temperature 0 & Low & \textbullet{} &  &  & \textbullet{} & \textbullet{} & \textbullet{} &  & 2025 Eaton Fire\\
Pixtral-12B & mistral-community/pixtral-12b & 2024-09-11 & Local & Greedy & None &  & \textbullet{} &  & \textbullet{} & \textbullet{} &  & \textbullet{} & None\\
Qwen2.5-VL-7B & Qwen/Qwen2.5-VL-7B-Instruct & 2025-01-28 & Local & Greedy & None &  & \textbullet{} & \textbullet{} & \textbullet{} & \textbullet{} &  & \textbullet{} & None\\
Qwen3-VL-8B & Qwen/Qwen3-VL-8B-Instruct & 2025-10-15 & Local & Greedy & None &  & \textbullet{} & \textbullet{} & \textbullet{} & \textbullet{} & \textbullet{} & \textbullet{} & None\\
Qwen3-VL-30B-A3B & Qwen/Qwen3-VL-30B-A3B-Instruct & 2025-10-04 & Local & Greedy & None & \textbullet{} & \textbullet{} & \textbullet{} & \textbullet{} & \textbullet{} &  & \textbullet{} & None\\
Qwen3.8-27B & Qwen/Qwen3.8-27B & 2026-08-14 & Local & Greedy & Off & \textbullet{} & \textbullet{} & \textbullet{} & \textbullet{} & \textbullet{} & \textbullet{} &  & 2025 Eaton Fire\\
\addlinespace[3.2pt]\multicolumn{14}{@{}l}{\textbf{Proprietary models (13)}}\\[0.8pt]
Claude Haiku 5.5 & claude-haiku-5-5 & 2026-10-07 & API & Default sampling & Low & \textbullet{} &  &  & \textbullet{} &  & \textbullet{} &  & 2025 Eaton Fire,\newline 2025 Hurricane Melissa\\
Claude Opus 5.5 & claude-opus-5-5 & 2026-09-22 & API & Default sampling & Low & \textbullet{} &  &  & \textbullet{} &  & \textbullet{} &  & 2025 Eaton Fire,\newline 2025 Hurricane Melissa\\
Claude Sonnet 5 & claude-sonnet-5 & 2026-06-30 & API & Default sampling & Low & \textbullet{} &  &  & \textbullet{} &  & \textbullet{} & \textbullet{} & 2025 Eaton Fire,\newline 2025 Hurricane Melissa\\
Gemini 3.1 Pro & gemini-3.1-pro-preview & 2026-02-19 & API & Temperature 0 & Low & \textbullet{} &  &  & \textbullet{} & \textbullet{} & \textbullet{} & \textbullet{} & 2025 Eaton Fire\\
Gemini 3.5 Flash-Lite & gemini-3.5-flash-lite & 2026-07-21 & API & Default sampling & Low & \textbullet{} &  &  & \textbullet{} &  & \textbullet{} &  & 2025 Eaton Fire\\
Gemini 3.8 Flash & gemini-3.8-flash & 2026-09-02 & API & Default sampling & Low & \textbullet{} &  &  & \textbullet{} &  & \textbullet{} &  & 2025 Eaton Fire\\
GPT-4o (May 2024) & gpt-4o-2024-05-13 & 2024-05-13 & API & Temperature 0 & None &  &  &  & \textbullet{} & \textbullet{} & \textbullet{} & \textbullet{} & None\\
GPT-5.4 & gpt-5.4 (snapshot 2026-03-05) & 2026-03-05 & API & Default sampling & Low & \textbullet{} &  &  & \textbullet{} &  & \textbullet{} & \textbullet{} & 2025 Eaton Fire\\
GPT-6 Luna & gpt-6-luna & 2026-09-22 & API & Default sampling & Low & \textbullet{} &  &  & \textbullet{} &  & \textbullet{} &  & 2025 Eaton Fire,\newline 2025 Hurricane Melissa\\
GPT-6.1 Sol & gpt-6.1-sol & 2026-09-29 & API & Default sampling & Low & \textbullet{} &  &  & \textbullet{} &  & \textbullet{} &  & 2025 Eaton Fire,\newline 2025 Hurricane Melissa\\
Grok 4.7 & grok-4.7 & 2026-09-21 & API & Temperature 0 & Low & \textbullet{} &  &  & \textbullet{} & \textbullet{} & \textbullet{} &  & 2025 Eaton Fire,\newline 2025 Hurricane Melissa\\
Muse Spark 1.3 & muse-spark-1.3 & 2026-09-02 & API & Temperature 0 & Low & \textbullet{} &  &  & \textbullet{} & \textbullet{} & \textbullet{} &  & 2025 Eaton Fire,\newline 2025 Hurricane Melissa\\
Qwen3.8-Max & qwen3.8-max-0902 & 2026-09-02 & API & Temperature 0 & Low & \textbullet{} &  &  & \textbullet{} & \textbullet{} & \textbullet{} &  & 2025 Eaton Fire,\newline 2025 Hurricane Melissa\\
\end{longtable}\par\vspace{3pt}

\clearpage
\noindent{\fontsize{9}{11}\selectfont\textbf{Supplementary Table 3 \textbar{} Models (continued)}}\par\vspace{3pt}
\noindent\textbf{b}, Other models.\par\vspace{2pt}
\begin{longtable}{@{}>{\raggedright\arraybackslash}p{28mm}>{\raggedright\arraybackslash}p{62mm}>{\raggedright\arraybackslash}p{34mm}@{}}
\toprule
{\fontsize{6}{7.2}\selectfont Model} & {\fontsize{6}{7.2}\selectfont Identifier} & {\fontsize{6}{7.2}\selectfont Used in}\\
\midrule
\endfirsthead
\multicolumn{3}{@{}l}{\textit{Supplementary Table 3b, continued}}\\[2pt]
\toprule
{\fontsize{6}{7.2}\selectfont Model} & {\fontsize{6}{7.2}\selectfont Identifier} & {\fontsize{6}{7.2}\selectfont Used in}\\
\midrule
\endhead
\bottomrule
\endlastfoot
OLMo-2 7B & allenai/OLMo-2-1124-7B-Instruct & Fig. 4b\\
OLMo-2 13B & allenai/OLMo-2-1124-13B-Instruct & Fig. 4b\\
\addlinespace[3.2pt]GeoChat & MBZUAI/geochat-7B & Fig. 7c\\
SkySenseGPT & ll-13/SkySenseGPT-7B-clip-lora & Fig. 7c\\
TEOChat & jirvin16/TEOChat & Fig. 7c\\
LLaVA-1.5 & liuhaotian/llava-v1.5-7b & Fig. 7c\\
Video-LLaVA & LanguageBind/Video-LLaVA-7B & Fig. 7c\\
\addlinespace[3.2pt]Qwen2-VL-2B & Qwen/Qwen2-VL-2B-Instruct & Extended Data Fig. 4c\\
Qwen2.5-VL-3B & Qwen/Qwen2.5-VL-3B-Instruct & Extended Data Fig. 4c\\
Qwen3-VL-4B & Qwen/Qwen3-VL-4B-Instruct & Extended Data Fig. 4c\\
Qwen2-VL-7B & Qwen/Qwen2-VL-7B-Instruct & Extended Data Fig. 4c\\
Qwen3-VL-32B & Qwen/Qwen3-VL-32B-Instruct & Extended Data Fig. 4c\\
InternVL3-2B & OpenGVLab/InternVL3-2B & Extended Data Fig. 4c\\
InternVL3-14B & OpenGVLab/InternVL3-14B & Extended Data Fig. 4c\\
InternVL3-38B & OpenGVLab/InternVL3-38B & Extended Data Fig. 4c\\
Gemma-3-12B & google/gemma-3-12b-it & Extended Data Fig. 4c\\
\end{longtable}\par\vspace{3pt}
\endgroup
\end{landscape}
 
\clearpage
\restoregeometry

\phantomsection\label{si:st4}\pdfbookmark[0]{Supplementary Table 4 | Coding of answers}{bm-st4}%
\begingroup\fontsize{8}{9.8}\selectfont\setlength{\parskip}{0pt}\setlength{\tabcolsep}{3pt}\renewcommand{\arraystretch}{1.0}%
\setlength{\heavyrulewidth}{0.75pt}\setlength{\lightrulewidth}{0.5pt}\setlength{\aboverulesep}{1.6pt}\setlength{\belowrulesep}{2.2pt}%
\setlength{\LTpre}{0pt}\setlength{\LTpost}{0pt}\setlength{\LTleft}{0pt}\setlength{\LTright}{0pt plus 1fill}%
\noindent{\fontsize{9}{11}\selectfont\textbf{Supplementary Table 4 \textbar{} Coding of answers}}\par\vspace{3pt}
\noindent Replies were coded by the rules below, which complete the rules stated in Methods.\par\vspace{3pt}
\begin{longtable}{@{}>{\raggedright\arraybackslash}p{36mm}>{\raggedright\arraybackslash}p{118mm}@{}}
\toprule
{\fontsize{7}{8.4}\selectfont Output} & {\fontsize{7}{8.4}\selectfont Answer and exclusions}\\
\midrule
\endfirsthead
\multicolumn{2}{@{}l}{\textit{Supplementary Table 4, continued}}\\[2pt]
\toprule
{\fontsize{7}{8.4}\selectfont Output} & {\fontsize{7}{8.4}\selectfont Answer and exclusions}\\
\midrule
\endhead
\bottomrule
\endlastfoot
Yes or no & Where the first line containing only `yes' or `no' differed from the first whole word `yes' or `no' in the final output, that line counted. The reasoning trace was the text before the model's end-of-reasoning marker, such as `\textless{}/think\textgreater{}'. For the models queried through an API, a reply was a refusal if it contained a refusal phrase, such as `I'm sorry', `I am unable' or `as an AI', or if a content filter stopped it, even when it also contained `yes' or `no'. For the locally run models, a refusal phrase did not make a reply a refusal. A reply stating that no image had been provided, or that the image was blank or only noise, was no answer unless it went on to answer the question, so the `no' of such a statement did not count.\\
\addlinespace[3pt]Percentage estimates & A valid number lay between 0 and 100, so years and counts above 100 were passed over. A range such as `10 to 20\%' counted at its lower end, and a number given after a stated decline, as in `if I had to estimate, 5', was an estimate. \\
\addlinespace[3pt]Damage boxes & A list with a formatting defect, such as a stray comma or a misspelt key, was coded through its complete boxes. Of two lists in one reply the last counted, and a single box given outside a list counted as a list of one. A reply that said in words that no building was damaged, such as `There are no damaged buildings', was an empty list, whereas a reply that declined without a list was no answer. A reply cut at the output limit kept its complete boxes, as did a list that opened more boxes than it completed. An answer was invalid when more than half of its boxes fell outside the image or, with at least three boxes, when more than half repeated one box or stepped across the image at a constant size and spacing. A list whose coordinates all lay between 0 and 2 was a placeholder and invalid, unless the model used a 0 to 1 scale. A box covering at least half of the image counted as a box but not in the test of where box centres fell. Coordinates were interpreted in each model's own order and unit (pixels of the image as the model received it, a 0 to 1 scale or a 0 to 1,000 grid), fixed from its no-cue answers on damaged images. Grok~4.7 and \mbox{GLM-5.3-Flash} mixed units, so each of their answers was interpreted by the range of its values. Values all at most 1 meant a 0 to 1 scale, values all at most 100 a 0 to 100 scale and any value above the side of the image a 0 to 1,000 grid; other answers were in pixels.\\
\addlinespace[3pt]Response decisions & The answer was taken from the first non-empty line. In the single-site decision, it was the option named first on that line, or the option named first anywhere in the reply when the line named none. In the two-site decision, it was the letter that opened the line, also written as `Site A' or in bold, or else the one site that the line named. A first line that named both sites or neither gave no answer. \\
\addlinespace[3pt]Text-only checks of the control places & Main comparison: any reply without `NO KNOWLEDGE' named an event. Response decisions: a reply named a disaster unless it denied knowing one, and disasters dated only before 1990 were not counted.\\
\addlinespace[3pt]Event-knowledge recall & A date answer was correct when the month and year of onset stood in one sentence, at most 15 characters apart.\\
\addlinespace[3pt]Recognition of recent disasters & Keywords for the place, country, year and hazard were matched anywhere in a reply; a reply that opened with `no knowledge', or contained it in fewer than 80 characters, was a denial.\\
\addlinespace[3pt]Inferred locations & A match counted only as a whole word, with or without an adjectival ending (for example, Iraqi). Negated mentions and the regions Central, Latin, North and South America, the Americas and the Indian Ocean were skipped. A country name or adjective that formed part of the name of another place was coded as the country in which that place lies (for example, Port of Spain as Trinidad and Tobago). If that place spans several countries, such as the Gulf of Mexico, or the name is shared by places in different countries, such as Panama City, the name was skipped. Country adjectives describing a building that the same sentence placed in another named place were also skipped. After a skipped name, the next place named counted. A country's own longer name, a country whose name contains another country's name (Papua New Guinea, South Sudan) and descriptions that are not names (northern Mexico, the Libyan coast) were coded as the country they name. \\
\addlinespace[3pt]Yes or no in the safeguard conditions & The `yes' or `no' opening the first line, unless a labelled final answer gave another; after a structured description, the last line giving `yes' or `no'. The same rules coded self-critique, the agents and every self-consistency sample.\\
\addlinespace[3pt]`Unsure' & A first line opening with `unsure' or an equivalent statement, such as `uncertain', `cannot tell', `unclear' or `insufficient information'.\\
\addlinespace[3pt]Agent replies & The first reply without a tool call, or the fourth reply if every reply called the tool.\\
\addlinespace[3pt]Withheld answers & The stated confidence was the number on the second line, or one next to `yes', `no' or `confidence'. A response reporting damage without a confidence was not withheld. Under self-consistency, one sample coded `unsure', refused or unreadable made the case an abstention.\\
\end{longtable}\par\vspace{3pt}
\noindent The keyword and place-name lists and the full parser are in the released data and code.\par
\endgroup
 
\clearpage

\phantomsection\label{si:st5}\pdfbookmark[0]{Supplementary Table 5 | Sites, labels and control places of the response decisions}{bm-st5}%
\begingroup\fontsize{8}{9.8}\selectfont\setlength{\parskip}{0pt}\setlength{\tabcolsep}{3pt}\renewcommand{\arraystretch}{1.0}%
\setlength{\heavyrulewidth}{0.75pt}\setlength{\lightrulewidth}{0.5pt}\setlength{\aboverulesep}{1.6pt}\setlength{\belowrulesep}{2.2pt}%
\setlength{\LTpre}{0pt}\setlength{\LTpost}{0pt}\setlength{\LTleft}{0pt}\setlength{\LTright}{0pt plus 1fill}%
\noindent{\fontsize{9}{11}\selectfont\textbf{Supplementary Table 5 \textbar{} Sites, labels and control places of the response decisions}}\par\vspace{3pt}
\begin{longtable}{@{}>{\raggedright\arraybackslash}p{36mm}>{\raggedright\arraybackslash}p{118mm}@{}}
\toprule
{\fontsize{7}{8.4}\selectfont Element} & {\fontsize{7}{8.4}\selectfont Rule}\\
\midrule
\endfirsthead
\multicolumn{2}{@{}l}{\textit{Supplementary Table 5, continued}}\\[2pt]
\toprule
{\fontsize{7}{8.4}\selectfont Element} & {\fontsize{7}{8.4}\selectfont Rule}\\
\midrule
\endhead
\bottomrule
\endlastfoot
Sites & The intact sites were the 137 post-event images of intact areas, and the damaged sites the 144 post-event corpus images of the same disasters, collections and, where recorded, acquisitions with at least 10\% of their labelled buildings damaged.\\
\addlinespace[3pt]Pairs of the two-site decision & Each intact site was paired with two damaged sites of its disaster, each damaged site used about equally often (for the 2023 Marrakesh–Safi earthquake, its one damaged site). This gave 208 pairs in the 21 disasters that kept a control place. \\
\addlinespace[3pt]Sector labels & The sector label was `sector' followed by a letter drawn with a fixed seed from C to Z without I and O. The letters A and B were left out because they name the two sites of the two-site decision. In the single-site decision each site received one letter, and in the two-site decision each pair received two different letters, which stayed with their sites when the positions were exchanged. \\
\addlinespace[3pt]Control places & One per disaster (Supplementary Table~2): a place matched to the disaster's place name in country or region, settlement type and size, with its own Wikipedia article and an unambiguous name, written in the same format. No disaster of the corpus and Wikipedia and news searches found none from 1990 to 2025 that killed ten or more people, destroyed hundreds of buildings or drew sustained international news coverage. \\
\end{longtable}\par\vspace{3pt}
\endgroup
 
\clearpage
\newgeometry{left=15mm,right=15mm,top=25mm,bottom=25mm}
\setlength{\parskip}{0pt}
\begin{landscape}
\phantomsection\label{si:st6}\pdfbookmark[0]{Supplementary Table 6 | Countries of the residential images and their coding}{bm-st6}%
\begingroup\fontsize{7.5}{9.2}\selectfont\setlength{\tabcolsep}{3pt}\renewcommand{\arraystretch}{1.0}\setlength{\LTpre}{0pt}\setlength{\LTpost}{0pt}
\noindent{\fontsize{9}{11}\selectfont\textbf{Supplementary Table 6 \textbar{} Countries of the residential images and their coding}}\par\vspace{3pt}
\setlength{\LTleft}{0pt}\setlength{\LTright}{0pt}
\begin{longtable}{@{\extracolsep{\fill}}lccrrrcccrl@{}}
\toprule
 & & & \multicolumn{1}{c}{Peak annual deaths in} & \multicolumn{1}{c}{Deaths in} & \multicolumn{1}{c}{Conflict} & & & & & \\
 & ISO & Conflict- & \multicolumn{1}{c}{state-based armed conflict,} & \multicolumn{1}{c}{organized violence,} & \multicolumn{1}{c}{phrases in} & Arid & & Global & \multicolumn{1}{c}{GDP per head} & Country-cue set\\
Country & code & affected & \multicolumn{1}{c}{2014 to 2023 (year)} & \multicolumn{1}{c}{1989 to 2024} & \multicolumn{1}{c}{Dolma 1.7} & climate & Coastline & South & \multicolumn{1}{c}{(US\$)} & (Fig.~6c)\\
\midrule
\endfirsthead
\multicolumn{11}{@{}l}{\textit{Supplementary Table 6, continued}}\\[2pt]
\toprule
 & & & \multicolumn{1}{c}{Peak annual deaths in} & \multicolumn{1}{c}{Deaths in} & \multicolumn{1}{c}{Conflict} & & & & & \\
 & ISO & Conflict- & \multicolumn{1}{c}{state-based armed conflict,} & \multicolumn{1}{c}{organized violence,} & \multicolumn{1}{c}{phrases in} & Arid & & Global & \multicolumn{1}{c}{GDP per head} & Country-cue set\\
Country & code & affected & \multicolumn{1}{c}{2014 to 2023 (year)} & \multicolumn{1}{c}{1989 to 2024} & \multicolumn{1}{c}{Dolma 1.7} & climate & Coastline & South & \multicolumn{1}{c}{(US\$)} & (Fig.~6c)\\
\midrule
\endhead
\bottomrule
\endfoot
\bottomrule
\endlastfoot
\multicolumn{11}{@{}l}{\textbf{Countries of the residential images}}\\
Algeria & DZA & No & <\,1,000 & 21,247 & 10,506 & Yes & Yes & Yes & 6,051 & --\\
Australia & AUS & No & <\,1,000 & 2 & 9,327 & No & Yes & No & 65,130 & Not conflict-affected\\
Belarus & BLR & No & <\,1,000 & 0 & 723 & No & No & No & 10,279 & --\\
Belgium & BEL & No & <\,1,000 & 40 & 3,155 & No & Yes & No & 60,750 & Not conflict-affected\\
Brazil & BRA & No & <\,1,000 & 20,008 & 2,645 & No & Yes & Yes & 10,713 & --\\
Canada & CAN & No & <\,1,000 & 93 & 19,146 & No & Yes & No & 55,698 & Not conflict-affected\\
Estonia & EST & No & <\,1,000 & 0 & 391 & No & Yes & No & 34,418 & Not conflict-affected\\
Germany & DEU & No & <\,1,000 & 23 & 25,963 & No & Yes & No & 60,496 & --\\
Indonesia & IDN & No & <\,1,000 & 9,271 & 3,143 & No & Yes & Yes & 5,060 & --\\
Iran & IRN & No & <\,1,000 & 2,805 & 32,192 & No & Yes & Yes & 3,924 & --\\
Iraq & IRQ & Yes & 13,168 (2014) & 127,241 & 1,984,053 & Yes & Yes & Yes & 5,410 & Conflict-affected\\
Kazakhstan & KAZ & No & <\,1,000 & 0 & 492 & No & No & Yes & 14,692 & --\\
Kenya & KEN & No & <\,1,000 & 6,645 & 2,424 & No & Yes & Yes & 2,363 & --\\
Latvia & LVA & No & <\,1,000 & 0 & 312 & No & Yes & No & 26,312 & --\\
Libya & LBY & Yes & 1,742 (2016) & 14,003 & 61,003 & Yes & Yes & Yes & 6,449 & Conflict-affected\\
Lithuania & LTU & No & <\,1,000 & 0 & 585 & No & Yes & No & 32,959 & --\\
Moldova & MDA & No & <\,1,000 & 585 & 422 & No & No & No & 8,622 & --\\
Nicaragua & NIC & No & <\,1,000 & 565 & 6,129 & No & Yes & Yes & 3,173 & --\\
Nigeria & NGA & Yes & 4,637 (2015) & 73,822 & 10,826 & No & Yes & Yes & 1,224 & --\\
North Korea & PRK & No & <\,1,000 & 0 & 7,675 & No & Yes & Yes & -- & --\\
Philippines & PHL & Yes & 1,627 (2017) & 21,513 & 9,161 & No & Yes & Yes & 4,171 & --\\
Poland & POL & No & <\,1,000 & 2 & 5,088 & No & Yes & No & 28,420 & Not conflict-affected\\
Russia & RUS & No & <\,1,000 & 33,700 & 37,933 & No & Yes & No & 17,547 & --\\
Senegal & SEN & No & <\,1,000 & 2,195 & 147 & No & Yes & Yes & 1,955 & --\\
Serbia & SRB & No & <\,1,000 & 8,773 & 2,824 & No & No & No & 15,262 & --\\
Somalia & SOM & Yes & 3,862 (2023) & 64,575 & 21,210 & Yes & Yes & Yes & 661 & Conflict-affected\\
Syria & SYR & Yes & 66,071 (2014) & 413,311 & 389,904 & No & Yes & Yes & 1,057 & Conflict-affected\\
Tanzania & TZA & No & <\,1,000 & 155 & 566 & No & Yes & Yes & 1,319 & --\\
Trinidad and Tobago & TTO & No & <\,1,000 & 39 & 152 & No & Yes & Yes & 18,967 & --\\
Turkey & TUR & Yes & 1,445 (2016) & 32,084 & 8,088 & No & Yes & Yes & 18,599 & --\\
Ukraine & UKR & Yes & 91,485 (2022) & 245,068 & 1,218,636 & No & Yes & No & 5,866 & --\\
United Kingdom & GBR & No & <\,1,000 & 333 & 9,202 & No & Yes & No & 57,602 & --\\
United States & USA & No & <\,1,000 & 3,016 & 31,636 & No & Yes & No & 90,027 & --\\
Yemen & YEM & Yes & 23,366 (2021) & 67,779 & 187,515 & Yes & Yes & Yes & 634 & Conflict-affected\\
\pagebreak
\multicolumn{11}{@{}l}{\textbf{Other conflict-affected countries}}\\
Afghanistan & AFG & Yes & 35,787 (2021) & -- & -- & -- & -- & -- & -- & --\\
Azerbaijan & AZE & Yes & 7,636 (2020) & -- & -- & -- & -- & -- & -- & --\\
Burkina Faso & BFA & Yes & 4,608 (2023) & -- & -- & -- & -- & -- & -- & --\\
DR Congo & COD & Yes & 3,718 (2017) & -- & -- & -- & -- & -- & -- & --\\
Ethiopia & ETH & Yes & 163,210 (2022) & -- & -- & -- & -- & -- & -- & --\\
Israel & ISR & Yes & 26,722 (2023) & -- & -- & -- & -- & -- & -- & --\\
Mali & MLI & Yes & 1,440 (2022) & -- & -- & -- & -- & -- & -- & --\\
Myanmar & MMR & Yes & 2,453 (2023) & -- & -- & -- & -- & -- & -- & --\\
Pakistan & PAK & Yes & 3,173 (2014) & -- & -- & -- & -- & -- & -- & --\\
South Sudan & SSD & Yes & 1,674 (2014) & -- & -- & -- & -- & -- & -- & --\\
Sudan & SDN & Yes & 5,353 (2023) & -- & -- & -- & -- & -- & -- & --\\
\end{longtable}\par\vspace{3pt}
\noindent Deaths are best estimates from the UCDP Georeferenced Event Dataset, version 25.1, summed over each country's events; organized violence comprises state-based conflict, non-state conflict and one-sided violence. The coding as conflict-affected follows the definition in Methods.\par
\noindent Inferred locations in Gaza and the West Bank were assigned to Israel, under which UCDP records them.\par
\noindent Conflict phrases are the summed infini-gram counts in Dolma 1.7 of \textit{war in X}, \textit{X war}, \textit{X conflict}, \textit{bombing in X}, \textit{X civil war}, \textit{airstrikes in X} and \textit{fighting in X}, with X the country's name (Britain, America and Trinidad for the United Kingdom, the United States and Trinidad and Tobago). In Extended Data Fig.~5b, conflict intensity is the natural logarithm of 1 plus the deaths in organized violence and conflict salience the base-10 logarithm of 1 plus the phrase count, each standardized over the responses.\par
\noindent Arid climate, coastline and Global South were coded by hand (Methods); a coastline is a border on a sea or an ocean, the Caspian Sea counting as inland.\par
\noindent GDP per head is in current US dollars from the World Development Indicators of the World Bank (updated 1 July 2026), for 2025 or the latest year available (Syria 2022, Yemen 2018); there is no figure for North Korea. In Extended Data Fig.~5b, it is the control variable in the rows for the country of the image.\par
\noindent Country-cue set is the set from which the other-country cues of each image were drawn (Fig.~6c; Methods).\par
\endgroup
\end{landscape}

\begin{landscape}
\phantomsection\label{si:st7}\pdfbookmark[0]{Supplementary Table 7 | Safeguard conditions}{bm-st7}%
\begingroup\fontsize{7.5}{9.2}\selectfont\setlength{\tabcolsep}{2.6pt}\renewcommand{\arraystretch}{1.0}
\noindent{\fontsize{9}{11}\selectfont\textbf{Supplementary Table 7 \textbar{} Safeguard conditions}}\par\vspace{3pt}
\noindent\textbf{a}, All models.\par\vspace{2pt}
\noindent\begin{tabular*}{\linewidth}{@{\extracolsep{\fill}}>{\raggedright\arraybackslash}p{68mm}cccccccc@{}}
\toprule
 & & \multicolumn{1}{c}{Cue effect} & \multicolumn{1}{c}{Residual} & \multicolumn{1}{c}{Change in} & & \multicolumn{1}{c}{Change in false-alarm} & \multicolumn{1}{c}{Change in detection} & \multicolumn{1}{c}{Cases}\\
 & & \multicolumn{1}{c}{at baseline} & \multicolumn{1}{c}{cue effect,} & \multicolumn{1}{c}{cue effect,} & \multicolumn{1}{c}{Holm-} & \multicolumn{1}{c}{rate without a cue,} & \multicolumn{1}{c}{on damaged images,} & \multicolumn{1}{c}{answered}\\
Condition & Models & \multicolumn{1}{c}{(pp)} & \multicolumn{1}{c}{pp (95\% CI)} & \multicolumn{1}{c}{pp (95\% CI)} & \multicolumn{1}{c}{adjusted \textit{P}} & \multicolumn{1}{c}{pp (95\% CI)} & \multicolumn{1}{c}{pp (95\% CI)} & \multicolumn{1}{c}{(\%)}\\
\midrule
No safeguard (baseline) & 27 & -- & 14.5 (10.6 to 18.7) & -- & -- & -- & -- & 100.0\\
\addlinespace[2pt]\multicolumn{9}{@{}l}{\textbf{Prompting and reasoning}}\\
Evidence-only instruction & 27 & 14.5 & 3.3 (1.6 to 5.4) & −11.2 (−14.2 to −8.5) & ≤\,0.0015 & −5.1 (−6.7 to −3.7) & −21.3 (−24.5 to −18.2) & 100.0\\
`Unsure' option & 26 & 15.0 & 3.8 (2.1 to 5.8) & −11.2 (−13.9 to −8.3) & ≤\,0.0015 & −4.9 (−6.5 to −3.6) & −20.0 (−23.0 to −17.1) & 96.6\\
Length-matched control & 27 & 14.5 & 16.2 (11.9 to 20.9) & 1.6 (0.4 to 3.0) & 0.051 & −2.0 (−2.6 to −1.3) & −1.5 (−2.2 to −0.9) & 100.0\\
Structured description & 27 & 14.5 & 9.5 (6.1 to 12.9) & −5.0 (−6.3 to −3.6) & ≤\,0.0015 & 1.5 (0.6 to 2.5) & −4.8 (−6.2 to −3.4) & 99.5\\
Self-critique & 27 & 14.5 & 6.3 (3.9 to 8.8) & −8.2 (−10.2 to −6.2) & ≤\,0.0015 & −3.4 (−4.6 to −2.3) & −14.6 (−17.3 to −12.0) & 99.7\\
\addlinespace[2pt]\multicolumn{9}{@{}l}{\textbf{Sampling and filtering}}\\
Self-consistency & 19 & 13.1 & 13.0 (9.0 to 17.3) & −0.1 (−0.5 to 0.3) & 0.86 & −0.1 (−0.3 to 0.0) & 0.3 (0.0 to 0.6) & 82.7\\
Confidence filter, below 80 withheld & 27 & 14.5 & 12.2 (9.0 to 16.0) & −2.3 (−3.4 to −1.4) & ≤\,0.0015 & −1.7 (−2.3 to −1.2) & −3.6 (−4.5 to −2.8) & 93.8\\
\hspace*{1.2em}below 90 withheld & 27 & 14.5 & 8.4 (5.7 to 11.2) & −6.2 (−8.2 to −4.4) & ≤\,0.0015 & −4.7 (−6.4 to −3.4) & −12.4 (−14.8 to −10.0) & 82.2\\
\hspace*{1.2em}below 95 withheld & 27 & 14.5 & 4.0 (2.3 to 6.0) & −10.5 (−13.4 to −7.6) & ≤\,0.0015 & −6.5 (−8.7 to −4.5) & −20.8 (−24.4 to −17.0) & 73.8\\
Cross-model vote & 27 & 14.5 & 12.3 (5.2 to 20.6) & −2.3 (−6.9 to 3.6) & 0.86 & −6.5 (−8.5 to −4.7) & 3.7 (0.5 to 6.9) & 100.0\\
Consistency filter & 27 & 14.5 & 0 & −14.5 (−18.7 to −10.6) & -- & 2.3 (0.9 to 3.8) & 2.2 (1.1 to 3.3) & 83.5\\
\addlinespace[2pt]\multicolumn{9}{@{}l}{\textbf{External damage estimates, 27 models}}\\
SKAI reading in the prompt, both wordings & 27 & 14.5 & 8.2 (5.5 to 11.3) & −6.3 (−8.2 to −4.3) & ≤\,0.0015 & −1.3 (−3.2 to 0.8) & −3.6 (−6.1 to −1.4) & 100.0\\
\hspace*{1.2em}balanced & 27 & 14.5 & 6.7 (4.4 to 9.1) & −7.9 (−10.6 to −5.0) & -- & 0.8 (−1.9 to 4.1) & −1.1 (−4.4 to 2.1) & 100.0\\
\hspace*{1.2em}own judgment & 27 & 14.5 & 9.8 (6.5 to 13.9) & −4.7 (−6.3 to −3.1) & -- & −3.5 (−4.9 to −2.2) & −6.2 (−7.8 to −4.6) & 100.0\\
Microsoft workflow reading in the prompt, both wordings & 27 & 14.5 & 9.4 (6.6 to 12.5) & −5.1 (−7.0 to −3.2) & ≤\,0.0015 & −1.0 (−3.4 to 1.2) & −1.8 (−5.8 to 1.9) & 100.0\\
\hspace*{1.2em}balanced & 27 & 14.5 & 7.8 (5.4 to 10.3) & −6.7 (−9.2 to −4.3) & -- & 2.0 (−1.3 to 5.4) & 0.8 (−4.9 to 6.4) & 100.0\\
\hspace*{1.2em}own judgment & 27 & 14.5 & 11.0 (7.6 to 14.7) & −3.5 (−5.4 to −1.8) & -- & −4.1 (−5.8 to −2.5) & −4.5 (−7.0 to −2.0) & 100.0\\
Ground truth proportion (reference reading), both wordings & 27 & 14.5 & 6.6 (4.3 to 9.2) & −7.9 (−10.0 to −5.8) & ≤\,0.0015 & −5.9 (−7.9 to −4.3) & 3.8 (2.0 to 5.5) & 100.0\\
\hspace*{1.2em}balanced & 27 & 14.5 & 4.2 (2.8 to 6.0) & −10.3 (−13.3 to −7.5) & -- & −6.3 (−8.3 to −4.4) & 8.9 (6.4 to 11.4) & 100.0\\
\hspace*{1.2em}own judgment & 27 & 14.5 & 9.0 (5.9 to 12.7) & −5.5 (−7.2 to −3.8) & -- & −5.6 (−7.4 to −4.0) & −1.4 (−2.7 to 0.0) & 100.0\\
\addlinespace[2pt]\multicolumn{9}{@{}l}{\textbf{External damage estimates, 21 tool-capable models}}\\
SKAI reading in the prompt, both wordings & 21 & 16.2 & 9.4 (6.4 to 12.8) & −6.8 (−8.6 to −4.9) & ≤\,0.0015 & −0.9 (−2.1 to 0.5) & −4.5 (−6.3 to −2.8) & 100.0\\
\hspace*{1.2em}balanced & 21 & 16.2 & 8.2 (5.5 to 11.2) & −8.0 (−10.6 to −5.5) & -- & 0.4 (−1.5 to 2.7) & −2.3 (−5.1 to 0.4) & 100.0\\
\hspace*{1.2em}own judgment & 21 & 16.2 & 10.6 (7.2 to 14.7) & −5.5 (−7.1 to −3.9) & -- & −2.2 (−3.2 to −1.2) & −6.8 (−8.2 to −5.4) & 100.0\\
ReAct agent with SKAI, both instructions & 21 & 16.2 & 6.9 (4.6 to 9.6) & −9.2 (−12.0 to −6.8) & ≤\,0.0015 & 1.1 (−0.6 to 3.2) & −4.3 (−7.2 to −1.2) & 100.0\\
\hspace*{1.2em}balanced & 21 & 16.2 & 6.0 (3.7 to 8.6) & −10.2 (−13.4 to −7.3) & -- & 2.6 (−0.3 to 6.3) & −2.3 (−6.6 to 1.9) & 100.0\\
\hspace*{1.2em}own judgment & 21 & 16.2 & 7.8 (5.1 to 10.8) & −8.3 (−10.5 to −6.2) & -- & −0.3 (−1.4 to 0.6) & −6.3 (−8.4 to −4.1) & 100.0\\
\bottomrule
\end{tabular*}\par\vspace{3pt}
\noindent The baseline is the cue effect of the same models without the safeguard, on the cases the safeguard answered or, for the confidence filter, the consistency filter and the external damage estimates, on all cases.\par
\noindent Holm-adjusted P values come from one family of the 15 tested comparisons, adjusted over all models in part a and within each model group in part b (Methods). A value written ≤\,0.0015 is a bound set by the 10,000 random sign flips. The rows of a single wording or instruction and the consistency filter were not tested.\par
\noindent The SKAI reading and the Microsoft workflow reading were the shares of an image's labelled buildings that each detector counted as damaged or destroyed: SKAI at its released threshold (probability of damage at least 0.85), the workflow detector when at least half of a building's footprint was classed as damaged. The reference reading was the share labelled damaged or destroyed in the ground truth, 0.0 on intact images. Readings were given with one decimal place (prompts in Extended Data Table~3), and the agents' tool returned the SKAI reading.\par
\endgroup
\end{landscape}

\begin{landscape}
\begingroup\fontsize{7.5}{9.2}\selectfont\setlength{\tabcolsep}{2.6pt}\renewcommand{\arraystretch}{1.0}
\noindent{\fontsize{9}{11}\selectfont\textbf{Supplementary Table 7 \textbar{} Safeguard conditions (continued)}}\par\vspace{3pt}
\noindent\textbf{b}, Open-weight and proprietary models.\par\vspace{2pt}
\noindent\begin{tabular*}{\linewidth}{@{\extracolsep{\fill}}>{\raggedright\arraybackslash}p{68mm}cccccccc@{}}
\toprule
 & \multicolumn{4}{c}{Open-weight models} & \multicolumn{4}{c}{Proprietary models}\\
\cmidrule(lr){2-5}\cmidrule(l){6-9}
 & & \multicolumn{1}{c}{Residual cue effect,} & \multicolumn{1}{c}{Change in} & \multicolumn{1}{c}{Holm-} & & \multicolumn{1}{c}{Residual cue effect,} & \multicolumn{1}{c}{Change in} & \multicolumn{1}{c}{Holm-}\\
Condition & Models & \multicolumn{1}{c}{pp (95\% CI)} & \multicolumn{1}{c}{cue effect, pp} & \multicolumn{1}{c}{adjusted \textit{P}} & Models & \multicolumn{1}{c}{pp (95\% CI)} & \multicolumn{1}{c}{cue effect, pp} & \multicolumn{1}{c}{adjusted \textit{P}}\\
\midrule
No safeguard (baseline) & 14 & 14.4 (9.3 to 19.3) & -- & -- & 13 & 14.7 (10.5 to 19.3) & -- & --\\
\addlinespace[2pt]\multicolumn{9}{@{}l}{\textbf{Prompting and reasoning}}\\
Evidence-only instruction & 14 & 2.6 (1.0 to 4.5) & −11.7 & ≤\,0.0015 & 13 & 4.0 (1.9 to 6.3) & −10.7 & ≤\,0.0015\\
`Unsure' option & 13 & 3.3 (1.3 to 5.6) & −12.2 & ≤\,0.0015 & 13 & 4.3 (2.5 to 6.5) & −10.2 & ≤\,0.0015\\
Length-matched control & 14 & 16.0 (10.8 to 21.6) & 1.6 & 0.37 & 13 & 16.4 (12.0 to 21.2) & 1.7 & 0.10\\
Structured description & 14 & 9.8 (5.4 to 14.3) & −4.6 & 0.0024 & 13 & 9.2 (6.0 to 12.5) & −5.5 & ≤\,0.0015\\
Self-critique & 14 & 6.4 (3.1 to 9.8) & −7.9 & ≤\,0.0015 & 13 & 6.2 (3.8 to 8.7) & −8.6 & ≤\,0.0015\\
\addlinespace[2pt]\multicolumn{9}{@{}l}{\textbf{Sampling and filtering}}\\
Self-consistency (all five samples agree) & 14 & 13.4 (9.2 to 18.0) & −0.2 & 0.37 & 5 & 11.9 (6.7 to 17.8) & 0.1 & 0.89\\
Confidence filter, below 80 withheld & 14 & 13.6 (8.8 to 18.7) & −0.7 & 0.21 & 13 & 10.8 (7.7 to 14.1) & −4.0 & ≤\,0.0015\\
\hspace*{1.2em}below 90 withheld & 14 & 8.9 (4.9 to 13.2) & −5.5 & ≤\,0.0015 & 13 & 7.9 (5.5 to 10.1) & −6.8 & ≤\,0.0015\\
\hspace*{1.2em}below 95 withheld & 14 & 5.1 (2.2 to 8.5) & −9.3 & ≤\,0.0015 & 13 & 3.2 (1.9 to 4.5) & −11.5 & ≤\,0.0015\\
Cross-model vote & 14 & 14.0 (6.5 to 23.6) & 0.2 & 0.95 & 13 & 12.7 (5.9 to 20.2) & −2.0 & 0.81\\
Consistency filter & 14 & 0 & −14.4 & -- & 13 & 0 & −14.7 & --\\
\addlinespace[2pt]\multicolumn{9}{@{}l}{\textbf{External damage estimates, 27 models}}\\
SKAI reading in the prompt, both wordings & 14 & 9.3 (6.2 to 12.7) & −5.1 & 0.0048 & 13 & 7.1 (4.4 to 10.1) & −7.6 & ≤\,0.0015\\
\hspace*{1.2em}balanced & 14 & 7.1 (4.8 to 9.6) & −7.2 & -- & 13 & 6.2 (3.8 to 8.8) & −8.5 & --\\
\hspace*{1.2em}own judgment & 14 & 11.5 (7.3 to 16.5) & −2.9 & -- & 13 & 8.1 (5.0 to 11.8) & −6.6 & --\\
Microsoft workflow reading in the prompt, both wordings & 14 & 11.0 (7.6 to 14.5) & −3.4 & 0.080 & 13 & 7.7 (5.0 to 10.8) & −7.0 & ≤\,0.0015\\
\hspace*{1.2em}balanced & 14 & 8.2 (5.6 to 10.6) & −6.2 & -- & 13 & 7.5 (4.9 to 10.4) & −7.2 & --\\
\hspace*{1.2em}own judgment & 14 & 13.8 (9.4 to 18.5) & −0.6 & -- & 13 & 8.0 (4.9 to 11.8) & −6.7 & --\\
Ground truth proportion (reference reading), both wordings & 14 & 8.4 (5.5 to 11.5) & −6.0 & 0.0024 & 13 & 4.7 (2.6 to 7.1) & −10.0 & ≤\,0.0015\\
\hspace*{1.2em}balanced & 14 & 5.8 (3.9 to 7.9) & −8.6 & -- & 13 & 2.6 (1.3 to 4.2) & −12.1 & --\\
\hspace*{1.2em}own judgment & 14 & 11.0 (6.9 to 15.6) & −3.3 & -- & 13 & 6.8 (4.0 to 10.3) & −7.9 & --\\
\addlinespace[2pt]\multicolumn{9}{@{}l}{\textbf{External damage estimates, 21 tool-capable models}}\\
SKAI reading in the prompt, both wordings & 8 & 13.1 (8.8 to 17.8) & −5.4 & 0.0018 & 13 & 7.1 (4.4 to 10.1) & −7.6 & ≤\,0.0015\\
\hspace*{1.2em}balanced & 8 & 11.4 (7.8 to 15.2) & −7.1 & -- & 13 & 6.2 (3.8 to 9.0) & −8.5 & --\\
\hspace*{1.2em}own judgment & 8 & 14.8 (9.8 to 20.4) & −3.7 & -- & 13 & 8.1 (4.8 to 11.8) & −6.6 & --\\
ReAct agent with SKAI, both instructions & 8 & 5.9 (3.2 to 9.0) & −12.6 & ≤\,0.0015 & 13 & 7.5 (5.2 to 10.2) & −7.2 & ≤\,0.0015\\
\hspace*{1.2em}balanced & 8 & 5.0 (2.1 to 8.5) & −13.5 & -- & 13 & 6.6 (4.5 to 9.1) & −8.1 & --\\
\hspace*{1.2em}own judgment & 8 & 6.8 (4.0 to 10.1) & −11.7 & -- & 13 & 8.4 (5.6 to 11.5) & −6.3 & --\\
\bottomrule
\end{tabular*}\par\vspace{3pt}
\noindent Notes as in part a. \par
\endgroup
\end{landscape}